\documentclass[lettersize,journal]{IEEEtran}
\usepackage{amsmath,amsfonts}
\usepackage{algorithmic}
\usepackage{algorithm}
\usepackage{array}
\usepackage[caption=false,font=normalsize,labelfont=sf,textfont=sf]{subfig}
\usepackage{textcomp}
\usepackage{stfloats}
\usepackage{url}
\usepackage{verbatim}
\usepackage{graphicx}
\usepackage{cite}
\usepackage{booktabs}
\usepackage[table]{xcolor}
\usepackage{multirow}
\usepackage{graphicx}
\begin{document}
\bstctlcite{IEEEexample:BSTcontrol}

\title{AffordVLA: Injecting Affordance Representations into Vision-Language-Action Models via Implicit Feature Alignment}

\author{Weijie Kong$^{*}$,
        Zhian Su$^{*}$,
        Wei Yu,
        Huixu Dong$^{\dagger}$,~\IEEEmembership{Member,~IEEE}
\thanks{Authors are with Grasp Lab, School of Mechanical Engineering of Zhejiang University, Hangzhou 310027, China; Authors are also with Torch Kernel Co., Ltd., Hangzhou, 310000, P. R. China (†The corresponding author Huixu Dong, e-mail: huixudong@zju.edu.cn). }

\thanks{*Equal contribution. †Corresponding author.}
}




\maketitle

\begin{abstract}
Recent advances in Vision-Language-Action (VLA) models have shown strong potential for general-purpose robotic manipulation. However, the visual representations of most VLA models are often dominated by global object appearance and struggle to focus on task-relevant functional interaction regions, which limits their robustness in unstructured environments. Existing affordance-based methods typically rely on explicit mask injection or external perception modules, requiring additional annotations while introducing cascading perception errors and inference overhead. To address these limitations, we propose AffordVLA, an affordance-enhanced VLA framework that internalizes manipulation-centric affordance perception into VLA visual representations through implicit representation alignment. Specifically, we construct a zero-shot affordance teacher to extract task-conditioned affordance visual representations from RGB observations and language instructions. AffordVLA aligns the intermediate visual representations of the VLA with the affordance visual representations extracted by the teacher, thereby implicitly injecting manipulation-centric affordance perception into VLA visual representations and improving action accuracy. Extensive simulation and real-world experiments demonstrate that AffordVLA and its affordance teacher achieve state-of-the-art performance and outperform strong baselines. Ablation analyses show that AffordVLA effectively reshapes VLA visual representations while preserving inference efficiency, leading to improved manipulation success rates and training efficiency.

\end{abstract}


\section{Introduction}
\IEEEPARstart{R}{obotic} general-purpose manipulation in unstructured environments remains a central challenge in robot learning~\cite{dong-2}, and understanding object affordances is essential for addressing this challenge~\cite{3dapnet}. Unlike other visual properties of objects such as color and shape, affordances describe task-conditioned functional mappings, where the same object may expose different interaction regions under different task goals~\cite{uad,affordancenet}. For example, a knife affords grasping at its handle, whereas it affords cutting at its blade. Similar to human manipulation, where attention is directed toward task-relevant functional regions rather than entire objects, affordance representations provide a manipulation-centric form of visual representation~\cite{omnimanip,a0,manipvqa}. Such representations are critical for robust manipulation in unstructured environments~\cite{mcr,tars,sa-dem}. However, how to effectively inject affordance representations into robot manipulation policies remains a key open problem.

\begin{figure}[t]
    \centering
    \includegraphics[width=\columnwidth]{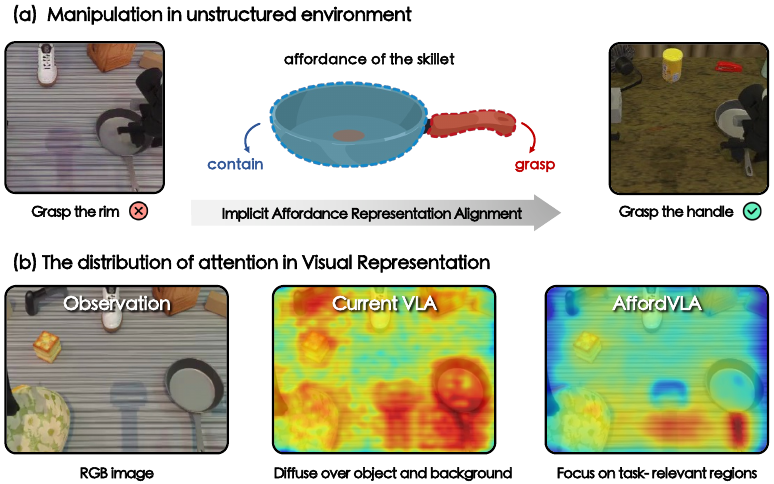}
    \caption{\textbf{Analysis of VLA visual representations in unstructured environments.} (a) Although the robot recognizes the skillet, it may grasp the body rather than the handle due to the lack of affordance-aware perception. (b) Current VLAs distribute visual attention over the entire object and background, whereas AffordVLA focuses on task-relevant functional interaction regions.}
    \label{fig:VLA visual representations}
\end{figure}

Recent Vision-Language-Action (VLA) models~\cite{rt-1,rt-2,pi_0} leverage pretrained Vision-Language Model (VLM) backbones to map visual observations and language instructions to robot actions in an end-to-end manner, demonstrating substantial potential as general-purpose manipulation policies~\cite{openvla,gr00t-n1}. However, current VLA models often suffer significant performance degradation when deployed in unstructured environments with complex textures or distracting objects~\cite{vla-jepa}. As illustrated in Fig.~1(a), even when the target object is correctly identified, the generated actions may still interact with functionally irrelevant regions.

To investigate the cause of these failures, we analyze the visual representations of VLA models to identify a fundamental limitation. As shown in Fig.~1(b), current VLA models tend to distribute visual attention across objects and even background regions, rather than concentrating it on functional interaction regions relevant to the current task. This phenomenon largely stems from VLM pretraining objectives, which emphasizes global semantics and object appearance~\cite{reconvla,sf}. Consequently, current VLA visual representations remain largely confined to global semantics and lack the affordance-aware perceptual capability required to map task intent onto functional object regions, thereby limiting manipulation robustness in unstructured environments~\cite{manipvqa,rt-affordance}.

To bridge the gap, recent works have begun to incorporate affordance information into robot manipulation policies. One line of work~\cite{rt-affordance,moka} follows an explicit injection paradigm, where affordance information produced by external detection models is directly fed into the policy network, demonstrating the effectiveness of affordance representations for improving manipulation robustness. However, several inherent limitations constrain their scalability and practicality as general-purpose solutions. First, these methods depend on large-scale affordance annotations, whereas existing open-vocabulary affordance datasets contain only tens of thousands of samples, limiting their scalability~\cite{kbag-net}. Second, the policy network relies on the affordance detection module at inference time, so its effectiveness is bounded by the detector's performance~\cite{u-sst}. Third, these additional modules introduce non-negligible inference overhead, making it difficult to satisfy the high-frequency, real-time control demands of robotic manipulation~\cite{vla-survey}. Implicit injection methods~\cite{coa-vla,palm} reduce the reliance on explicit inputs through latent query vectors. However, they typically focus on affordance-guided reasoning and planning rather than manipulation-centric affordance perception, while still requiring substantial affordance annotations and additional inference overhead. This raises an essential question: \textbf{\textit{can manipulation-centric affordance perception be implicitly internalized into VLA visual representations using only existing robot demonstration data, thereby eliminating the need for explicit affordance signals or external modules at inference time?}}

To address this question, we introduce AffordVLA, a framework that implicitly injects affordance information into VLA visual representations. First, we construct a zero-shot affordance teacher to generate normalized affordance representations as supervisory signals. Second, during training, we align the intermediate visual representations of the VLA with the affordance representations extracted by the teacher, thereby implicitly eliciting affordance-aware perception within the VLM backbone. Notably, AffordVLA can be trained directly on existing robot manipulation data without requiring additional annotations. Extensive simulation and real-world experiments demonstrate that AffordVLA significantly and consistently improves manipulation robustness in unstructured scenarios. On the RoboTwin benchmark, AffordVLA achieves state-of-the-art performance, surpassing the previous best baseline by 20.5\% in the Easy setting and 12.8\% in the Hard setting. Real-world experiments further validate the precise manipulation capability of AffordVLA in unstructured environments. Our contributions are summarized as follows:
\begin{itemize}
    \item We introduce AffordVLA, an affordance-enhanced VLA framework that internalizes manipulation-centric affordance information into VLA visual representations. AffordVLA introduces affordance supervision only during training and incurs zero additional inference overhead, thereby improving manipulation robustness while maintaining efficient inference.

    \item We design a zero-shot affordance teacher that generates task-conditioned affordance supervision without requiring task-specific affordance annotations, providing affordance supervision signals for VLA training.

    \item We propose an implicit affordance representation alignment mechanism that aligns the intermediate visual representations of the VLA with the affordance visual representations generated by the teacher. This mechanism enables the model to focus on task-relevant functional interaction regions while preserving the semantic representation capability of the original VLA.

    \item The simulation and real-world experiments demonstrate that AffordVLA achieves superior performance and higher data efficiency than existing baselines, particularly in unstructured scenarios.
\end{itemize}

\begin{figure*}[t]
    \centering
    \includegraphics[width=\textwidth]{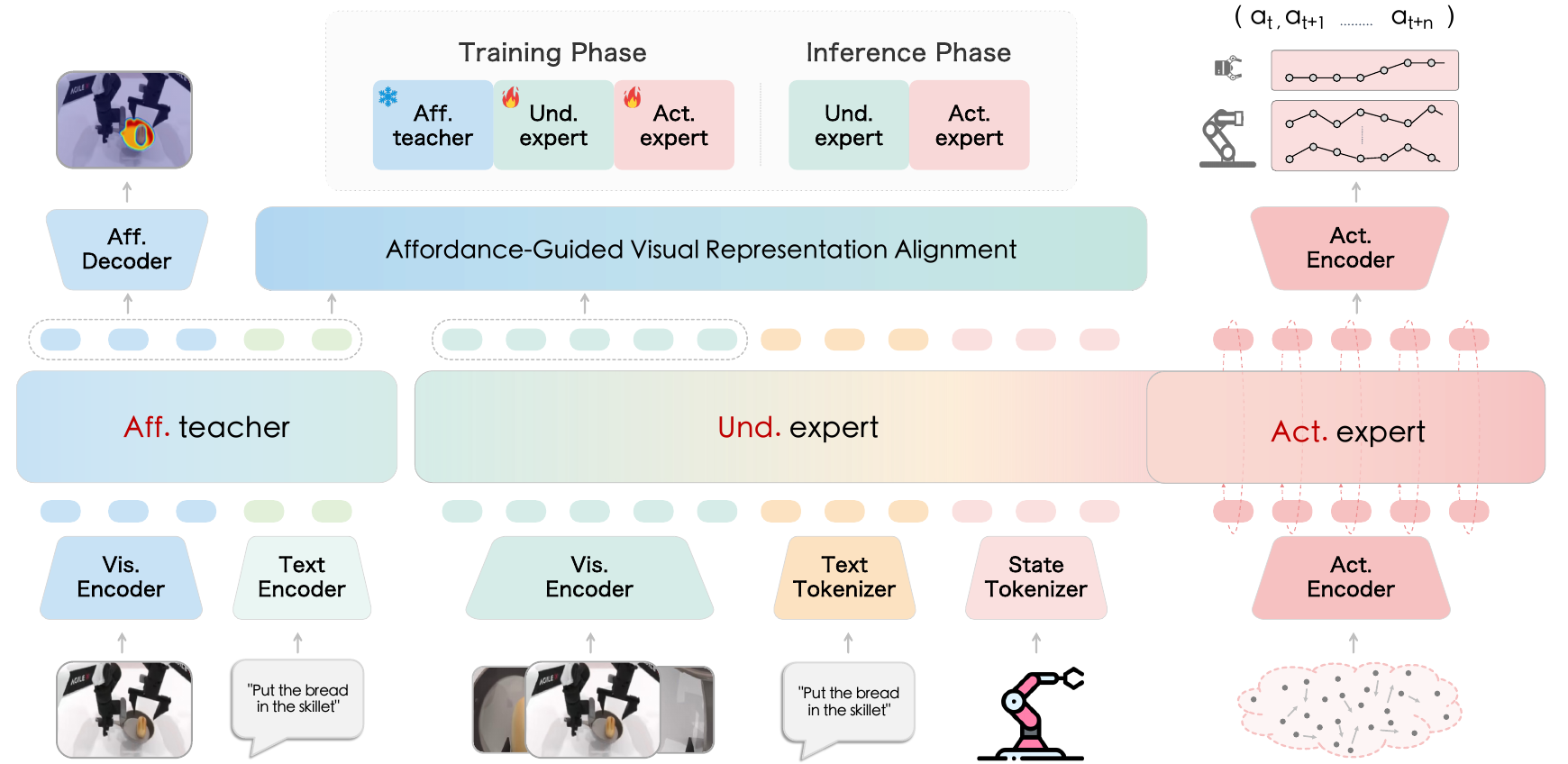} 
    \caption{\textbf{Overall architecture of AffordVLA.} The model consists of three components: an affordance teacher, an understanding expert, and an action expert. During training, the frozen affordance teacher provides task-conditioned affordance visual representations to supervise the intermediate visual representations of the understanding expert through representation alignment. During inference, the affordance teacher is removed, and only the understanding expert and action expert are retained for action generation.}
    \label{fig:affordvla_framework}
\end{figure*}

\section{Related work}
\subsection{Vision-Language-Action Models}
With the rapid progress of Multimodal Large Language Models (MLLMs)~\cite{gpt-4,qwen3}, VLA models have emerged as general-purpose robotic manipulation policies by integrating pretrained VLMs with robot action prediction modules. Existing works~\cite{rt-1,rt-2,openvla,pi_0,gr00t-n1} leverage the semantic knowledge acquired during VLM pretraining and combine it with diffusion models or Transformers to achieve end-to-end mapping from visual observations and language instructions to robot actions. Recent work has further expanded the capabilities of VLA models along several directions, including data scaling~\cite{pi_05,oxe}, enhanced reasoning~\cite{cot-vla,lare-vla}, visual prediction~\cite{f1,vla-jepa}, and reinforcement learning-based post-training~\cite{pi_06*,ig-rft,vla-rl}. However, most existing VLA works primarily focus on architecture design and data scaling, while their visual representations often lack manipulation-centric perceptual capabilities, leading to substantial drops in success rates in unstructured environments~\cite{mcr,manipvqa,vla-survey}. In contrast, our work implicitly aligns external affordance representations with VLA visual representations, endowing the VLA with manipulation-centric perceptual capabilities and achieving more robust manipulation in unstructured environments.

\subsection{Affordance Learning for Robotic Manipulation}
Affordance is commonly defined as the action possibilities offered by objects in the environment to an actor~\cite{Gibson}. In robotics, affordances are typically instantiated as task-relevant functional interaction regions on objects~\cite{affordancenet}. Early works~\cite{affordancenet,affpose,affordancellm,relanet} learn object affordances from manually annotated closed-vocabulary datasets and perform affordance detection in the form of masks or heatmaps. These works mainly demonstrate the benefits of affordances in limited manipulation settings~\cite{acpl} but remain difficult to deploy directly in closed-loop manipulation systems~\cite{clover}.

Recent works have further explored how to integrate affordance information into manipulation policy networks. One line of work~\cite{moka,uad,a0,manipgpt,rt-affordance} adopts an explicit injection paradigm, where affordance information, such as masks, heatmaps, or keypoints, is generated by external affordance detectors or multimodal large language models and then fed into the policy network, effectively improving manipulation robustness in unstructured environments. However, these methods typically depend on large-scale affordance annotations~\cite{kbag-net,a0}. Their inference-time performance is also bounded by the accuracy of external modules~\cite{u-sst}, while these modules introduce substantial latency~\cite{manipgpt,a0}. Another line of work~\cite{coa-vla,palm} attempts to inject affordance information implicitly through latent query vectors, thereby reducing the reliance on explicit inputs. However, these methods typically focus on affordance-guided reasoning and planning rather than manipulation-centric affordance perception, while still requiring substantial affordance annotations and additional inference overhead~\cite{mcr,lare-vla}. Unlike these approaches, we construct a zero-shot affordance teacher that provides task-conditioned affordance supervision without task-specific annotations. We further internalize affordance perception into VLA visual representations through implicit representation alignment, improving manipulation performance while incurring zero additional overhead at inference time.

\subsection{Representation Supervision}
Model representations can be supervised through reconstruction or alignment to improve their internal quality~\cite{sf,reconvla,r3m}. This representation supervision paradigm has been shown to effectively improve downstream task performance~\cite{repa,3drs}. Reconstruction-based supervision improves visual representations by imposing denoising reconstruction objectives in the latent space~\cite{genhancer}. Existing works~\cite{reconvla,ross} supervise visual representations by reconstructing input images or target manipulation regions with denoising architectures. However, such methods struggle to filter out redundant background details and visual noise~\cite{vla-jepa}, and are prone to conflict with the semantic representations acquired during VLM pretraining~\cite{cma-vla}. This makes them less suitable for guiding VLAs to learn manipulation-relevant visual representations.

Alignment-based methods instead align intermediate model representations with external reference representations. Recent works have adopted this paradigm by aligning intermediate VLA visual representations with 3D geometry-aware representations~\cite{sf,spatialvla} or future observation embeddings~\cite{flare}, achieving improvements in both performance and efficiency. However, their alignment targets mainly capture general semantic or spatial features and lack manipulation-centric perceptual information. Our method follows the alignment supervision paradigm but introduces affordance representations as alignment targets for VLA training. With supervision from the affordance teacher, our method injects manipulation-centric affordance perception into VLA visual representations while preserving the original semantic capabilities of the VLA.

\section{Methodology}
To address the misalignment between VLA visual representations and task-relevant manipulation cues, we propose AffordVLA, an affordance-enhanced vision-language-action framework based on implicit representation alignment. The key idea is to introduce external affordance supervision during training and internalize affordance priors into intermediate visual representations of the VLA through feature alignment, thereby steering VLA visual representations away from diffuse attention over global appearance cues and toward task-relevant functional interaction regions. Specifically, Sec.~III-A presents the overall architecture of AffordVLA, Sec.~III-B describes the zero-shot affordance teacher, Sec.~III-C introduces the implicit affordance representation alignment strategy, and Sec.~III-D provides the implementation and training details.

\subsection{Architecture Overview}
As illustrated in Fig.~2, AffordVLA consists of three core components: the affordance teacher, the understanding expert, and the action expert. These components are responsible for generating affordance supervision, modeling multimodal semantics, and producing robot actions, respectively.

\textbf{Affordance Teacher.} This component serves as an external visual supervisor during training, providing high-quality task-conditioned affordance visual representations for the VLA backbone. Given a language instruction and the current RGB observation, it extracts normalized affordance representations that highlight manipulation-relevant functional regions while suppressing non-essential object parts and background distractions. The detailed design of the affordance teacher is presented in Sec.~III-B.

\textbf{Understanding Expert.} This component serves as the semantic reasoning core of AffordVLA. It builds on the VLM backbone of $\pi 0.5$~\cite{pi_05}, using Gemma-2B as the LLM backbone and SigLIP-So400m as the visual encoder. The module processes three synchronized inputs: the language instruction $l_t$, image observation $I_t$, and robot state $s_t$. The language instruction is first tokenized into text tokens and mapped into the feature space through the VLM token embedding layer. The image observation is encoded into visual tokens by the visual encoder. The robot state is discretized and incorporated into the model together with the text tokens. These multimodal features are then fed into the VLM Transformer backbone for joint modeling, producing a unified contextual condition for subsequent action generation:

\begin{equation}
C_t = F_{\mathrm{VLM}}\!\left([V_t, L_t, S_t]\right),
\end{equation}
where $V_t = E_{\mathrm{vis}}(I_t)$, $L_t = T_{\mathrm{txt}}(l_t)$, and $S_t = T_{\mathrm{state}}(s_t)$.

\textbf{Action Expert.} This component transforms the multimodal context from the understanding expert into continuous robot control commands. Following $\pi 0$~\cite{pi_0}, we build a lightweight Transformer and use conditional flow matching~\cite{flow-matching} to model the continuous distribution of a future action chunk $A_t = [a_t, a_{t+1}, \ldots, a_{t+H-1}]$. Formally, the action expert takes the context condition, a noisy action chunk, and the flow timestep $\tau$ as inputs, and outputs the corresponding vector-field prediction. During training, the action expert is randomly initialized and optimized with a conditional flow-matching objective. Given a noisy action chunk $A_t^{\tau} = \tau A_t + (1-\tau)\epsilon$, where $\tau \sim U(0,1)$ is the flow timestep and $\epsilon \sim \mathcal{N}(0,I)$ is Gaussian noise, the action expert predicts the vector field $v_{\theta}(A_t^{\tau}, C_t)$. It is optimized by minimizing the following action loss:

\begin{equation}
\mathcal{L}_{\mathrm{action}} =
\mathbb{E}\!\left[
\left\| v_{\theta}(A_t^{\tau}, C_t) - (A_t - \epsilon) \right\|_2^2
\right].
\end{equation}

To better capture temporal coupling within the action chunk, the action expert applies a fully bidirectional attention mask over all action tokens, allowing them to attend to one another.

\subsection{Zero-Shot Affordance Teacher}
To inject affordance perception into VLA visual representations, we first construct an external affordance teacher that provides priors over task-relevant functional regions. Since robotic manipulation is inherently open-world, with diverse manipulated objects and different functional regions even within the same object under different tasks. Therefore, perception models that rely on large-scale manually annotated affordance labels are difficult to scale and cannot adequately cover the diversity of real robotic tasks and scenes. To address this issue, we introduce a zero-shot affordance teacher that generates task-conditioned affordance supervision without additional annotations, training, or fine-tuning.

\begin{figure}[t]
    \centering
    \includegraphics[width=\columnwidth]{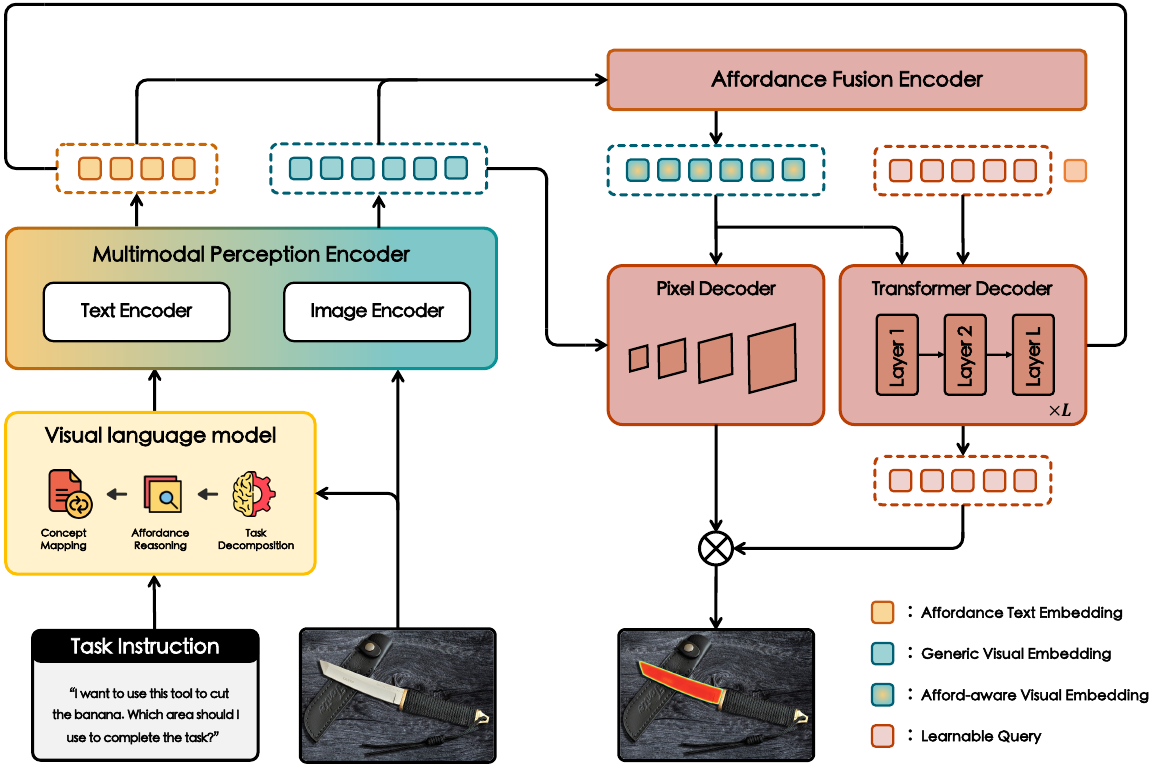}
    \caption{\textbf{Overall architecture of the zero-shot affordance teacher.} Given an RGB observation and a language instruction, the task parsing module first generates affordance concept prompts. The open-vocabulary affordance perception module then extracts task-conditioned affordance visual representations and produces pixel-level affordance predictions.}
    \label{fig:aff_teacher}
\end{figure}

The overall architecture of the zero-shot affordance teacher is shown in Fig.~3. It consists of two components: a multimodal task parsing module and an open-vocabulary affordance perception module. Given the current observation image $I_t$ and task instruction $l_t$, the multimodal task parsing module first interprets high-level task semantics. Specifically, we employ Qwen3-VL~\cite{qwen3} to transform the abstract manipulation objective into part-level visual concept prompts for the current task:

\begin{equation}
p_t = R(I_t, l_t),
\end{equation}
where $R(\cdot)$ denotes the task parsing module, and $p_t$ denotes the generated part-level affordance concept prompt. For example, given the instruction ``hammer the nail with the hammer'', the parsing module does not produce an object-level description. Instead, it outputs fine-grained concept prompts associated with the current manipulation stage, such as ``hammer handle'' or ``hammer head''.

Then, the parsed concept prompt $p_t$ and observation image $I_t$ are fed into the open-vocabulary affordance perception module. The concept prompt and image are encoded by the text encoder and image encoder, respectively, producing affordance text embeddings and generic visual embeddings. Both encoders inherit the architecture and weights of the SAM3 perception encoder~\cite{sam3}, ensuring that visual and textual features share a semantic space. We further feed the generic visual embeddings and affordance text embeddings into an affordance fusion encoder, where affordance concepts are injected into visual features through multi-layer Transformer-based cross-modal interaction. In this way, generic visual embeddings are transformed into affordance-aware visual embeddings. Based on these embeddings, a multi-scale decoder and a Transformer decoder are used to produce pixel-level affordance predictions. The overall process of the open-vocabulary affordance perception module is formulated as:

\begin{equation}
\left(Z_t^{\mathrm{aff}}, M_t^{\mathrm{aff}}\right) = T_{\mathrm{aff}}(I_t, p_t),
\end{equation}
where $T_{\mathrm{aff}}(\cdot)$ denotes the open-vocabulary affordance perception module, $Z_t^{\mathrm{aff}}$ is the task-conditioned affordance visual representation, and $M_t^{\mathrm{aff}}$ is the corresponding pixel-level affordance prediction.

It is important to note that the supervision used for subsequent implicit representation alignment is not the final explicit affordance mask $M_t^{\mathrm{aff}}$, but the intermediate affordance visual representation $Z_t^{\mathrm{aff}}$. This is because $Z_t^{\mathrm{aff}}$ preserves both task-semantic constraints and local spatial structure during cross-modal interaction, making it a more suitable target for VLA representation alignment. In contrast, the pixel-level prediction $M_t^{\mathrm{aff}}$ is mainly used to regularize the teacher to produce more stable task-conditioned representations and to validate its zero-shot affordance perception capability.

\subsection{Implicit Affordance Representation Alignment}
To endow VLA visual representations with manipulation-centric affordance perception, we avoid concatenating explicit affordance outputs to the policy input. Instead, we internalize the affordance priors provided by the external teacher into the intermediate visual representations of the VLA through implicit representation alignment. Formally, we use the task-conditioned affordance visual representations produced by the zero-shot affordance teacher as alignment targets for the visual features of the understanding expert. This encourages the VLA backbone to focus on task-relevant affordance regions rather than relying only on global semantics or appearance, while keeping the original inference pipeline unchanged.

Specifically, given the current observation image $I_t$ and task instruction $l_t$, the zero-shot affordance teacher first produces a task-conditioned affordance representation ${Z_t^{\mathrm{aff}} \in \mathbb{R}^{N \times d}}$, where $N$ denotes the number of flattened spatial tokens and $d$ denotes the feature dimension. To preserve the spatial ordering of image tokens in the autoregressive VLA backbone, we further introduce a positional embedding $P \in \mathbb{R}^{N \times d}$ and construct the alignment target as:

\begin{equation}
\tilde{Z}_t^{\mathrm{aff}} = Z_t^{\mathrm{aff}} + P.
\end{equation}

Let the visual representation from the $m$-th layer of the understanding expert be $X_t^{V,(m)} \in \mathbb{R}^{N_v \times d_v},$ where $N_v$ is the number of visual tokens and $d_v$ is the feature dimension of the intermediate VLA features. Since the teacher and the understanding expert operate in different representation spaces, we first normalize $X_t^{V,(m)}$ and resize it to the teacher resolution using bilinear interpolation, before mapping it into the teacher feature space with a two-layer MLP.

\begin{equation}
\hat{X}_t^{V,(m)} = W_2\,\sigma\!\left(W_1\,\mathrm{Norm}\!\left(\mathrm{Resize}\!\left(X_t^{V,(m)}\right)\right)\right),
\end{equation}
where $\mathrm{Resize}(\cdot)$ denotes bilinear interpolation, $\mathrm{Norm}(\cdot)$ denotes the VLM normalization operation, and $\sigma(\cdot)$ denotes a nonlinear activation function. After this transformation, the aligned VLA features satisfy $\hat{X}_t^{V,(m)} \in \mathbb{R}^{N \times d}$ and become directly comparable to the teacher representations.

We then enforce feature alignment by maximizing the cosine similarity between the VLA visual representations and the teacher affordance representations. The affordance alignment loss is formulated as:

\begin{equation}
\mathcal{L}_{\mathrm{align}}
=
-\frac{1}{N}\sum_{i=1}^{N}
\cos\!\left(
\hat{x}_{t,i}^{V,(m)},
\tilde{z}_{t,i}^{\mathrm{aff}}
\right),
\end{equation}
where $\hat{x}^{V,(m)}_{t,i}$ and $\tilde{z}_{t,i}^{\mathrm{aff}}$ denote the aligned VLA feature and the teacher feature at spatial location $i$, respectively. By maximizing their consistency, the model is encouraged to attend to task-relevant functional interaction regions, thereby injecting affordance awareness into VLA visual representations implicitly.

Recent works have shown that not all visual layers in a VLA are equally suitable for representation supervision~\cite{sf,huang2024}. Shallow layers preserve more local texture details, but the effect of alignment at these layers can be gradually weakened by subsequent cross-modal modeling and may not be stably retained in the higher-level representations used for action generation. In contrast, very deep layers tend to converge toward a more abstract modality-agnostic semantic space, where vision-specific structural details are increasingly compressed. Therefore, these layers are also suboptimal for affordance supervision. Based on this trade-off, we apply alignment to an intermediate-deep layer that balances semantic abstraction with visual detail preservation.

The final training objective of AffordVLA combines the action generation loss with the affordance alignment loss. Since the action expert models future action chunks with conditional flow matching, the overall objective is 

\begin{equation}
\mathcal{L}_{\mathrm{AffordVLA}}
=
\mathcal{L}_{\mathrm{action}}
+
\lambda \mathcal{L}_{\mathrm{align}},
\end{equation}
where $\lambda$ denotes the weight of the alignment loss. Through joint optimization, the VLA preserves its original visual-semantic modeling ability while being shaped into an affordance-aware representation space sensitive to task-relevant functional regions. Moreover, the affordance teacher is used only during training and remains frozen throughout optimization. It is removed entirely at inference time. As a result, the aligned AffordVLA retains the same inference pipeline as a standard VLA without introducing additional structures or computational overhead, while implicitly acquiring task-conditioned affordance perception.

\subsection{Implementation Details}
\textbf{Model details.} The affordance teacher contains approximately 0.8B parameters. Its visual encoder takes high-resolution inputs of size $1008\times1008$, while both the affordance fusion encoder and the Transformer decoder consist of six Transformer layers. The understanding expert is inherited from PaliGemma-3B and contains approximately 3.0B parameters with 18 layers. The action expert adopts an 18-layer Transformer architecture with approximately 0.3B parameters. During inference, the number of flow-matching denoising steps is set to 10, and the action chunk horizon is set to $H=30$. In the implicit feature alignment module, we align the visual features from the 12th layer of the understanding expert (out of 18 layers, with $d_v=2048$) with the task-conditioned visual features from the final layer of the affordance teacher ($n=6$, with $d=256$). The alignment module includes bilinear interpolation, sinusoidal positional encoding, and VLM normalization.

\textbf{Training details.} The alignment loss weight is set to $\lambda=0.5$. All models are trained on a single NVIDIA A100-SXM4-80GB GPU with a global batch size of 32. We use AdamW as the optimizer with $\beta_1=0.9$ and $\beta_2=0.999$. The learning rate follows a warmup cosine decay schedule, increasing linearly to $1\times10^{-4}$ during the first 10,000 steps and then decaying to $5\times10^{-5}$ with cosine annealing. We enable bfloat16 mixed-precision training throughout and keep the affordance teacher frozen during the entire training process. For simulation experiments, the models are trained for 30,000 steps in total. For real-world experiments, the models are trained for 50 epochs. All inference experiments are conducted on a single NVIDIA A100-SXM4-40GB GPU.

\section{Experiments}
This section systematically evaluates the affordance perception capability, manipulation performance, and implicit representation alignment mechanism of AffordVLA. To this end, we conduct experiments to answer the following questions:
\begin{enumerate}
    \item Can the zero-shot affordance teacher generate high-quality task-conditioned affordance priors in open-world scenarios? (Sec.~IV-A)
    \item Can AffordVLA outperform existing policy models in simulation tasks under dynamic changes and strong visual distractions? (Sec.~IV-B)
    \item Can AffordVLA maintain high success rates and efficient inference in real-world robotic tasks, while outperforming explicit affordance injection? (Sec.~IV-C)
    \item Does implicit affordance representation alignment effectively improve the model's manipulation performance and visual representations? (Sec.~IV-D)
\end{enumerate}

\subsection{Task-Conditioned Affordance Prediction}
\textbf{Dataset.} To evaluate whether the zero-shot affordance teacher can provide reliable task-conditioned affordance priors in open-world scenarios, we primarily evaluate it on AGD20K~\cite{agd20k}, following LOCATE~\cite{locate}. AGD20K is a large-scale affordance benchmark with action and object labels, making it suitable for evaluating generalization to unseen object categories. Notably, AGD20K does not provide natural-language task instructions, but only predefined action and object labels. To match the task-conditioned input format of our affordance teacher, we follow the setting of UAD~\cite{uad} and convert the action and object labels of each sample into a natural-language task description using the template ``region to \textless action\textgreater{} of the \textless object\textgreater'', where \textless action\textgreater{} and \textless object\textgreater{} are taken from the original AGD20K annotations. During evaluation, we strictly follow the original Unseen split of AGD20K. Our zero-shot affordance teacher does not use any AGD20K training samples, nor does it require additional training, fine-tuning, or annotation adaptation. It is directly evaluated on the Unseen test set in a zero-shot manner.

\textbf{Metrics.} We adopt the standard evaluation metrics used on AGD20K, including Kullback-Leibler Divergence (KLD)~\cite{kld}, Similarity Metric (SIM)~\cite{sim}, and Normalized Scanpath Saliency (NSS)~\cite{nss}. KLD measures the divergence between the predicted affordance distribution and the ground-truth distribution, where a lower value indicates a closer match to the annotation. SIM measures the distributional overlap between the prediction and the ground truth, where a higher value indicates stronger similarity. NSS measures the response strength of the prediction over the ground-truth affordance regions, where a higher value indicates that the model better highlights key interaction regions. Unlike binary mask-based segmentation metrics, these metrics treat affordance prediction as a continuous response distribution over the image space, making them more suitable for evaluating whether the model can generate fine-grained, spatially continuous, and task-relevant functional region priors.

\textbf{Baselines.} We compare the proposed zero-shot affordance teacher with state-of-the-art baselines. To provide a comprehensive comparison across different supervision paradigms, we group the baselines into fully supervised, weakly supervised, and zero-shot methods. Fully supervised methods typically rely on pixel-level affordance annotations for training, including 3DOI~\cite{3doi}, AffordanceLLM~\cite{affordancellm}, OOAL~\cite{ooal}, and AffordanceSAM~\cite{affordancesam}, among others. Weakly supervised methods do not directly use full pixel-level affordance labels but learn functional regions from demonstrations, interactions, or other weak supervision signals, including InteractionHotspots~\cite{InteractionHotspots}, LOCATE~\cite{locate}, INTRA~\cite{intra}, and BiT-Align~\cite{BiT-Align}, among others. Zero-shot methods are not trained on AGD20K and mainly rely on the open-world understanding and visual grounding capabilities of large models, including LISA-7B~\cite{lisa}, M2SA-7B~\cite{mmsa}, Molmo+SAM~\cite{affogato}, Espresso-2D~\cite{affogato}, UAD~\cite{uad}, and the proposed zero-shot affordance teacher.

\textbf{Results.} The quantitative results are shown in Tab.~I. Our affordance teacher achieves 0.905 KLD, 0.496 SIM, and 1.906 NSS on the AGD20K Unseen split. Among all weakly supervised and zero-shot methods, our method achieves the best performance on all three metrics, showing that the teacher can generate high-quality task-conditioned affordance predictions without any additional training. Compared with UAD, our affordance teacher reduces KLD from 1.878 to 0.905 and improves NSS from 1.092 to 1.906, indicating that its prediction distribution is closer to the ground-truth annotations and produces stronger responses over key functional regions. Compared with fully supervised methods, our model still achieves the best results on KLD and NSS, demonstrating strong open-world generalization capability. Although its SIM score is slightly lower than the 0.503 achieved by the fully supervised Espresso-2D, our score of 0.496 is very close.

Fig.~4 shows qualitative results of our affordance teacher on the AGD20K Unseen split under the open-world generalization setting. The results show that, given different action semantics, the model can dynamically localize the corresponding functional interaction regions conditioned on the task description, rather than simply segmenting the entire object. For example, for tasks such as ``take photo'', ``type on'' and ``hold'', the predictions focus on local regions directly related to action execution and produce more fine-grained affordance distributions. The last two rows further show that, when facing objects such as bicycles, cups, and knives, the model predicts different affordance regions according to different task instructions. Both quantitative and qualitative results demonstrate that our zero-shot affordance teacher can handle diverse unseen object categories and task semantics without additional annotations or fine-tuning, providing reliable affordance priors for subsequent implicit representation alignment.

\subsection{Comparison on the Simulation Benchmark}

\begin{figure*}[t]
    \centering
    \includegraphics[width=\textwidth]{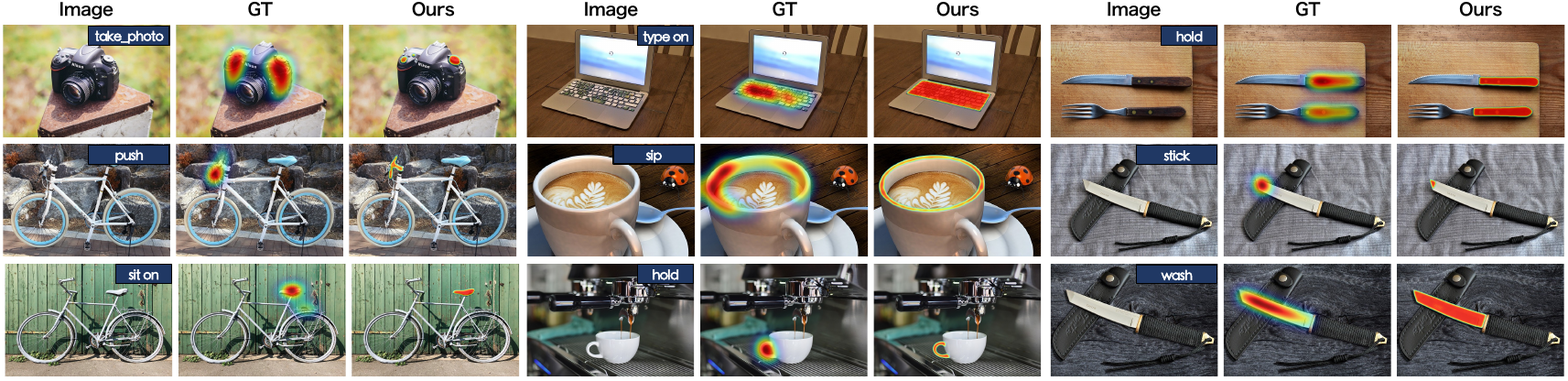}
    \caption{\textbf{Qualitative results of zero-shot affordance prediction on the AGD20K dataset.} The results show that the affordance teacher can predict task-relevant functional interaction regions in open-world human activity scenarios conditioned on different action semantics.}
    \label{fig:right_column_image}
\end{figure*}

\textbf{Simulation Environment.} To evaluate the performance and robustness of AffordVLA in simulated manipulation tasks, we conduct experiments on the RoboTwin2.0 benchmark~\cite{robotwin}. RoboTwin2.0 is a real-to-sim benchmark for bimanual robotic manipulation, providing high-fidelity physical interaction environments and supporting task evaluation under different difficulty levels. We select five representative tasks: Beat Block Hammer, Place Bread Skillet, Place Bread Basket, Place Can Basket, and Place Cans Plasticbox. RoboTwin2.0 provides two evaluation settings, Easy and Hard. The Easy setting mainly evaluates the basic manipulation capability of a model in relatively standard and in-distribution scenes. In contrast, the Hard setting introduces stronger domain randomization and visual distractions, including cluttered object layouts, texture variations, lighting changes, and more complex initial states, making it closer to real unstructured manipulation scenarios. Therefore, the Hard setting better reflects the model's capability to perceive task-relevant regions and remain robust to visual distractions.

\textbf{Baselines and Metric.} We compare AffordVLA with mainstream robotic policy models, including ACT~\cite{act}, Diffusion Policy~\cite{dp}, DP3~\cite{dp3}, RDT~\cite{rdt}, and $\pi_0$~\cite{pi_0}. ACT and DP are representative imitation learning policies, DP3 introduces 3D representations to improve spatial perception, and RDT and $\pi_0$ represent recent general-purpose robot policies based on VLA models. All experiments strictly follow the original RoboTwin2.0 evaluation protocol. For each task, each method is trained using only 50 disturbance-free Easy demonstrations, and the trained model is evaluated under both Easy and Hard test settings. AffordVLA follows the training configuration described in Sec.~III-D, while the remaining methods use the official training settings of RoboTwin2.0. We use success rate (SR) as the evaluation metric. For each task and difficulty setting, the model is evaluated over 100 independent trials, and the proportion of successful trials is reported.

\begin{table}[t]
\centering
\caption{\textbf{Quantitative results of affordance prediction on the AGD20K Unseen split.} Lower KLD indicates better performance, while higher SIM and NSS indicate better performance. The best and second-best results are highlighted in bold and underlined, respectively. Our method achieves STATE-OF-THE-ART performance on KLD and NSS, while obtaining the second-best result on SIM.}
\label{tab:agd20k_unseen}
\setlength{\tabcolsep}{4pt}
\renewcommand{\arraystretch}{1.08}
\resizebox{\columnwidth}{!}{
\begin{tabular}{llccc}
\toprule
Supervision & Method & KLD ($\downarrow$) & SIM ($\uparrow$) & NSS ($\uparrow$) \\
\midrule
\multirow{5}{*}{Fully}
& 3DOI\cite{3doi}              & 3.565 & 0.227 & 0.657 \\
& AffordanceLLM\cite{affordancellm}     & 1.463 & 0.377 & 1.070 \\
& OOAL\cite{ooal}              & 1.070 & 0.461 & 1.503 \\
& AffordanceSAM\cite{affordancesam}     & 1.271 & 0.486 & \underline{1.597} \\
& Espresso-2D\cite{affogato}       & \underline{1.034} & \textbf{0.503} & 1.550 \\
\midrule
\multirow{9}{*}{Weakly}
& InteractionHotspots\cite{InteractionHotspots} & 1.994 & 0.237 & 0.577 \\
& Cross-View-AG\cite{agd20k}       & 1.787 & 0.285 & 0.829 \\
& AffCorrs\cite{AffCorrs}            & 1.618 & 0.348 & 1.021 \\
& LOCATE\cite{locate}              & 1.405 & 0.372 & 1.157 \\
& INTRA\cite{intra}               & 1.365 & 0.375 & 1.209 \\
& WSMA\cite{WSMA}                & 1.335 & 0.382 & 1.220 \\
& WSAG-PLSP\cite{WSAG-PLSP}           & 1.153 & 0.437 & 1.418 \\
& R-Mamba\cite{r-mamba}             & 1.310 & 0.397 & 1.279 \\
& BiT-Align\cite{BiT-Align}           & 1.331 & 0.371 & 1.302 \\
\midrule
\multirow{6}{*}{Zero-shot}
& LISA-7B\cite{lisa}         & 1.830 & 0.256 & 0.765 \\
& M$^2$SA-7B\cite{mmsa}      & 1.926 & 0.227 & 0.657 \\
& Molmo + SAM\cite{affogato}     & 1.953 & 0.226 & 0.718 \\
& Espresso-2D\cite{affogato}     & 1.571 & 0.376 & 1.016 \\
& UAD\cite{uad}             & 1.878 & 0.407 & 1.092 \\
\rowcolor{gray!15}
& Ours       & \textbf{0.905} & \underline{0.496} & \textbf{1.906} \\
\bottomrule
\end{tabular}
}
\end{table}

\begin{table*}[t]
\centering
\caption{\textbf{Success rates (\%) on the RoboTwin2.0 simulation benchmark under Easy and Hard settings.} The best result in each column is shown in bold. AffordVLA achieves the best average success rate and shows stronger robustness.}
\label{tab:robotwin_transposed}
\setlength{\tabcolsep}{5pt}
\renewcommand{\arraystretch}{1.15}
\resizebox{\textwidth}{!}{
\begin{tabular}{p{1.2cm}cccccccccccc}
\toprule
\multirow{2}{*}{Method}
& \multicolumn{2}{c}{Beat Block Hammer}
& \multicolumn{2}{c}{Place Bread Skillet}
& \multicolumn{2}{c}{Place Bread Basket}
& \multicolumn{2}{c}{Place Can Basket}
& \multicolumn{2}{c}{Place Cans Plasticbox}
& \multicolumn{2}{c}{Average} \\
\cmidrule(lr){2-3}
\cmidrule(lr){4-5}
\cmidrule(lr){6-7}
\cmidrule(lr){8-9}
\cmidrule(lr){10-11}
\cmidrule(lr){12-13}
& Easy & Hard & Easy & Hard & Easy & Hard & Easy & Hard & Easy & Hard & Easy & Hard \\
\midrule
ACT\cite{act}        & 56.0          & 3.0           & 7.0  & 0.0  & 6.0  & 0.0  & 1.0  & 0.0  & 16.0 & 0.0  & 17.2 & 0.6 \\
DP\cite{dp}         & 42.0          & 0.0           & 11.0 & 0.0  & 14.0 & 0.0  & 18.0 & 0.0  & 40.0 & 0.0  & 25.0 & 0.0 \\
DP3\cite{dp3}        & 72.0          & 8.0           & 19.0 & 0.0  & 26.0 & 1.0  & \textbf{67.0} & 2.0  & 48.0 & 3.0  & 46.4 & 2.8 \\
RDT\cite{rdt}        & \textbf{77.0} & \textbf{37.0} & 5.0  & 1.0  & 10.0 & 2.0  & 19.0 & 6.0  & 6.0  & 5.0  & 23.4 & 10.2 \\
$\pi_0$\cite{pi_0}    & 43.0          & 21.0          & 23.0 & 1.0  & 17.0 & 4.0  & 41.0 & 5.0  & 34.0 & 2.0  & 31.6 & 6.6 \\
\rowcolor{gray!15}
Ours & 68.0          & 29.0          & \textbf{61.0} & \textbf{28.0} & \textbf{53.0} & \textbf{11.0} & 65.0 & \textbf{50.0} & \textbf{59.0} & \textbf{26.0} & \textbf{61.2} & \textbf{28.8} \\
\bottomrule
\end{tabular}
}
\end{table*}

\textbf{Results.} The results are shown in Tab.~II. AffordVLA achieves the best overall performance on RoboTwin2.0, with average success rates of 61.2\% and 28.8\% under the Easy and Hard settings, respectively. Compared with the strongest baselines, AffordVLA outperforms DP3 under the Easy setting by 14.8 percentage points and significantly surpasses RDT under the Hard setting by 18.6 percentage points. This indicates that AffordVLA not only achieves stronger manipulation performance in standard scenes but also shows better robustness under complex visual distractions and domain randomization.

From the task-level results, AffordVLA shows clear advantages in several challenging tasks. For example, in Place Can Basket, AffordVLA achieves a 50.0\% success rate under the Hard setting, substantially outperforming other methods. In Place Bread Skillet and Place Cans Plasticbox, AffordVLA also achieves 28.0\% and 26.0\% success rates under the Hard setting, respectively, while most baselines nearly fail on these tasks. Our observations suggest that, when test scenes contain lighting changes, complex backgrounds, or cluttered object distractors, the visual representations of other methods are more easily affected by irrelevant appearance cues, leading to unstable target localization and action execution failures. In contrast, the proposed implicit affordance representation alignment encourages the model to consistently attend to task-relevant functional regions, thereby reducing the impact of visual distractions on policy decisions and improving manipulation success rates. These results confirm the importance of affordance representation alignment for improving the robustness of VLAs in complex simulated manipulation scenarios.

\subsection{Comparison on the Real-World Tasks}
\begin{figure*}[t]
    \centering
    \includegraphics[width=\textwidth]{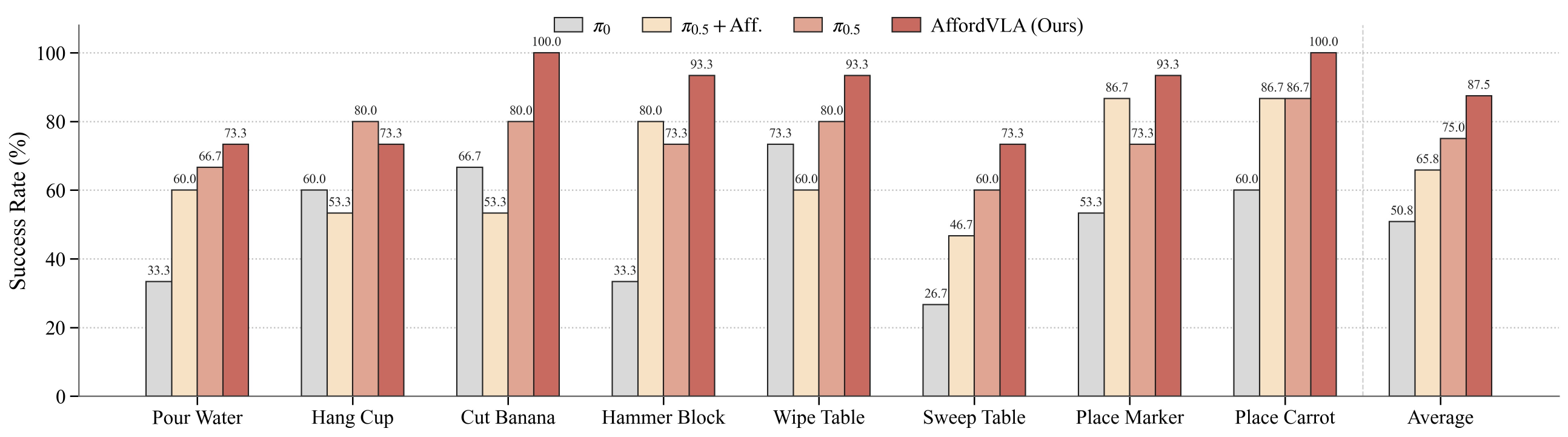} 
    \caption{\textbf{Results of real-world robotic experiments.} We conduct 15 trials for each model and report the average success rate. The results show that AffordVLA achieves better performance than other VLA models. $\pi_{0.5}$+Aff. denotes the method that explicitly injects affordance masks into $\pi_{0.5}$, further demonstrating that implicit affordance representation alignment improves real-world manipulation performance more effectively than explicit affordance injection.}
    \label{fig:vla_framework}
\end{figure*}

\textbf{Real-World Setup.} To further evaluate the manipulation capability of AffordVLA in real physical environments, we conduct systematic experiments on a real-world robotic platform. The real-world robotic platform is shown in Fig.~6. The platform consists of a UR5 robotic arm equipped with a Robotiq 2F-85 adaptive parallel gripper. The visual system includes a Kinect DK camera and a RealSense camera, which provide a third-person global view and awrist view, respectively. The model takes high-resolution RGB images as input. Policy inference is deployed on a server equipped with an NVIDIA A100-SXM4-40GB GPU, while the robot control client runs on a local PC. The server and local client are connected through socket-based communication. The local PC is equipped with an Intel Core i7-12700KF CPU, 32GB RAM, and an RTX 4070 GPU.

\begin{figure}[t]
    \centering
    \includegraphics[width=\columnwidth]{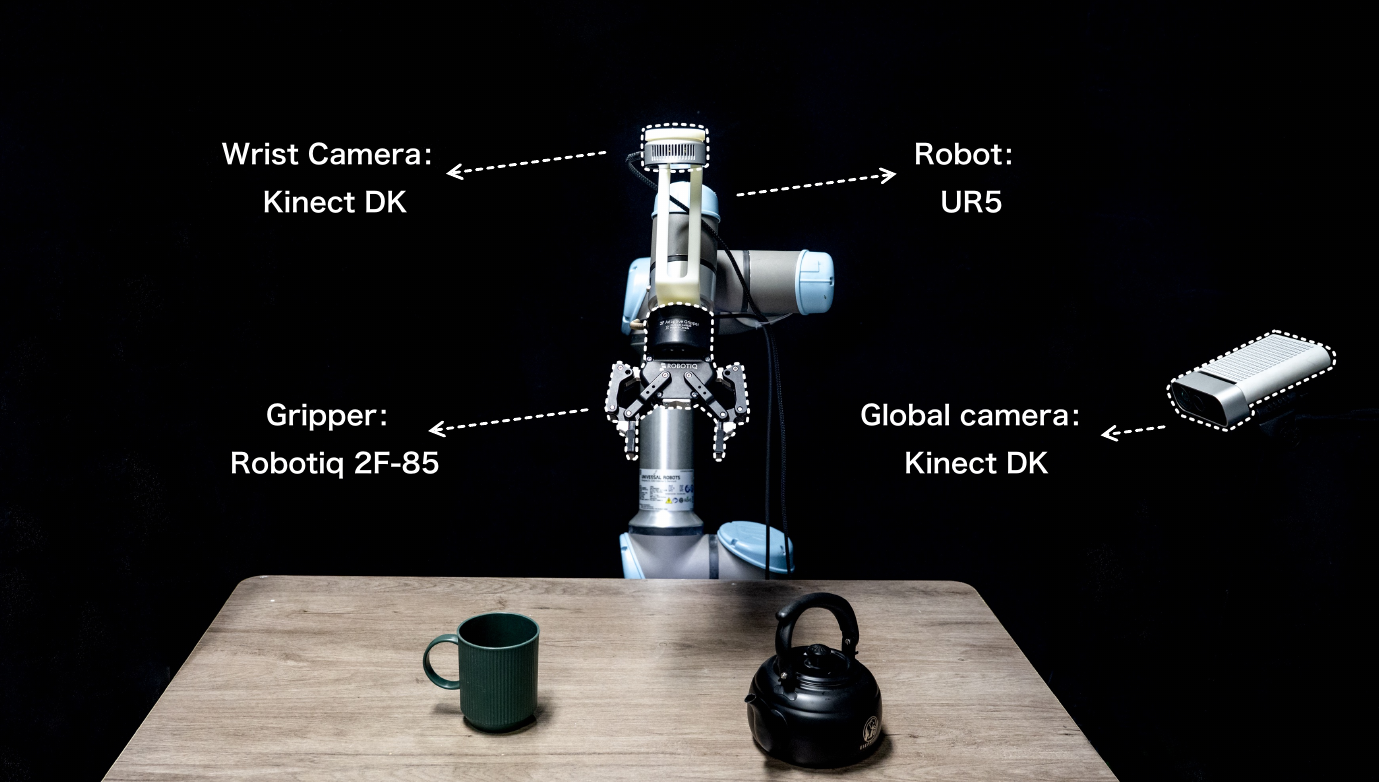}
    \caption{\textbf{Real-world robotic platform.}}
    \label{fig:Real-world robotic platform}
\end{figure}

\textbf{Tasks.} We design eight real-world manipulation tasks to evaluate the model's overall capability in rigid tool use, deformable object manipulation, container placement, and cluttered-scene sorting. These tasks include pouring water, hanging a mug, cutting a banana with a knife, striking a block with a hammer, sweeping a tabletop with a broom, wiping a stain with a cloth, placing a marker into a pen holder, and sorting objects in a cluttered scene. These tasks cover several representative challenges, including functional part localization, tool pose control, deformable contact, precise placement, and target selection under strong visual distractions.

\textbf{Baselines.} We compare AffordVLA with $\pi_0$~\cite{pi_0}, $\pi_{0.5}$~\cite{pi_05}, and $\pi_{0.5}$ + explicit affordance input. Here, $\pi_{0.5}$ + explicit affordance input denotes a variant that directly feeds the explicit affordance mask generated by the affordance teacher into $\pi_{0.5}$ as an additional visual input. This comparison is used to examine whether the proposed implicit representation alignment is more suitable for VLA policy learning than explicit affordance injection. All models are trained and evaluated under the same setting. Each task uses 50 collected real-world trajectories for training, and task success rate is adopted as the primary evaluation metric.

\begin{figure*}[t]
    \centering
    \includegraphics[width=\textwidth]{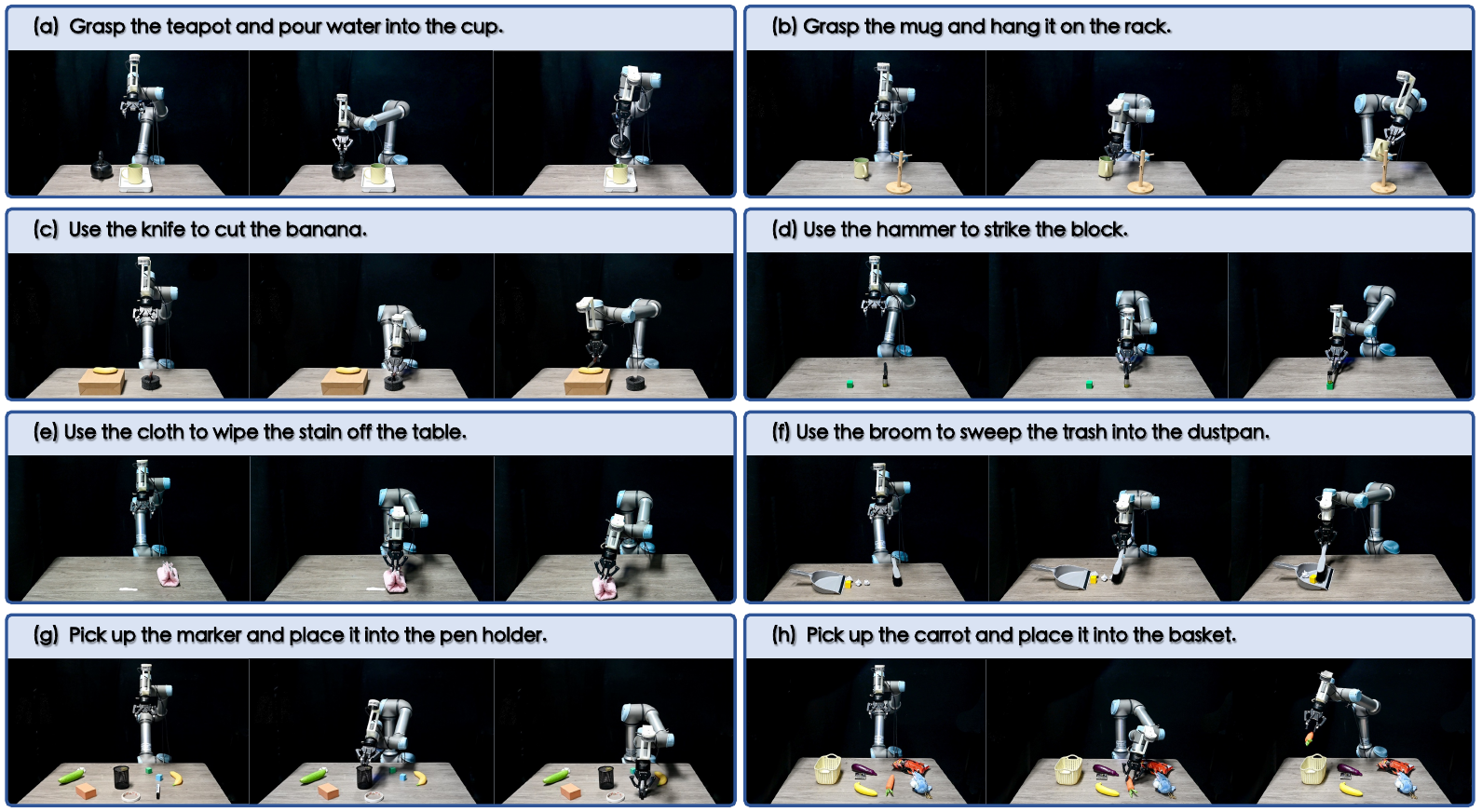} 
    \caption{\textbf{Execution examples of AffordVLA on real-world manipulation tasks.} The language instruction for each task is shown above the corresponding subfigure. The examples demonstrate diverse tasks, including pouring water, hanging a mug, cutting a banana, striking a block, sweeping, wiping, placing a marker, and sorting objects in a cluttered scene.}
    \label{fig:vla_framework}
\end{figure*}

\textbf{Results.} The real-world task results are shown in Fig.~5. Overall, AffordVLA achieves the best performance across the eight real-world robotic tasks, clearly outperforming both $\pi_0$ and $\pi_{0.5}$. This indicates that implicit affordance representation alignment enables the model to more consistently attend to task-relevant functional interaction regions in real physical environments, thereby improving manipulation success rates. In particular, for tool-use and cluttered-scene tasks, AffordVLA more accurately identifies key regions such as hammer handles, knife handles, broom handles, pen holder openings, and target objects, reducing failures caused by background distractions and attention to non-functional regions.

\begin{table}[t]
\centering
\caption{\textbf{Inference efficiency comparison in real-world experiments.} $\pi_{0.5}$ + explicit affordance input denotes the setting where the affordance detection mask is explicitly provided to $\pi_{0.5}$ as an additional visual input. The results show that, compared with explicit affordance injection, implicit representation alignment better preserves the inference efficiency of the robotic system.}
\label{tab:realworld_efficiency}
\setlength{\tabcolsep}{6pt}
\renewcommand{\arraystretch}{1.1}
\begin{tabular}{lcc}
\toprule
Method & Frequency (Hz) & Latency (ms) \\
\midrule
$\pi_{0.5}$\cite{pi_05} & 12.5 & 80.2 \\
$\pi_{0.5}$ + explicit affordance input   & 4.8 & 206.9 \\
\rowcolor{gray!15}
AffordVLA (Ours) & 12.4 & 80.4 \\
\bottomrule
\end{tabular}
\end{table}

It is worth noting that explicitly feeding affordance masks does not lead to stable improvements. Although this strategy can alleviate visual distractions in a few tasks, such as the cluttered Place Marker task where the affordance mask highlights the target region, it performs worse than the original $\pi_{0.5}$ in most tasks. This is because directly concatenating explicit masks to the VLA visual input may disrupt the original visual representation distribution of the pretrained VLM, making it difficult for the model to use the additional information consistently. In contrast, AffordVLA does not require explicit affordance inputs during inference. Instead, it internalizes affordance priors into intermediate visual representations during training, thereby obtaining more robust task-conditioned perception while preserving the original input format.

Fig.~7 shows the execution processes of AffordVLA on real-world tasks. The model can complete the closed-loop process from target recognition and functional region localization to manipulation execution based on the language instruction and current observation. For example, in the knife-cutting and hammer-striking tasks, the model needs to understand the functional roles of different tool parts. In sweeping, wiping, and cluttered sorting tasks, the model must suppress distractors in complex backgrounds and stably select the target region.

\textbf{Inference Efficiency.} In addition to manipulation success rates, we further compare the average inference efficiency of different methods in real-world tasks, as shown in Tab.~III. The original $\pi_{0.5}$ achieves an inference frequency of 12.5 Hz with an average latency of 80.2 ms. When explicit affordance masks are provided as input, the inference frequency drops to 4.8 Hz and the latency increases to 206.9 ms, indicating that explicit affordance injection introduces substantial computational overhead and struggles to meet the requirements of high-frequency real-time control. In contrast, AffordVLA reaches an inference frequency of 12.4 Hz with an average latency of 80.4 ms, maintaining an inference speed above 12 Hz, comparable to $\pi_{0.5}$. This shows that, compared with explicit affordance-mask input, implicit representation alignment not only brings more stable manipulation performance in real-world tasks but also enables more efficient inference, making it more suitable for deployment in real robotic systems.

\subsection{Ablation Studies}
\begin{figure*}[t]
    \centering
    \includegraphics[width=\textwidth]{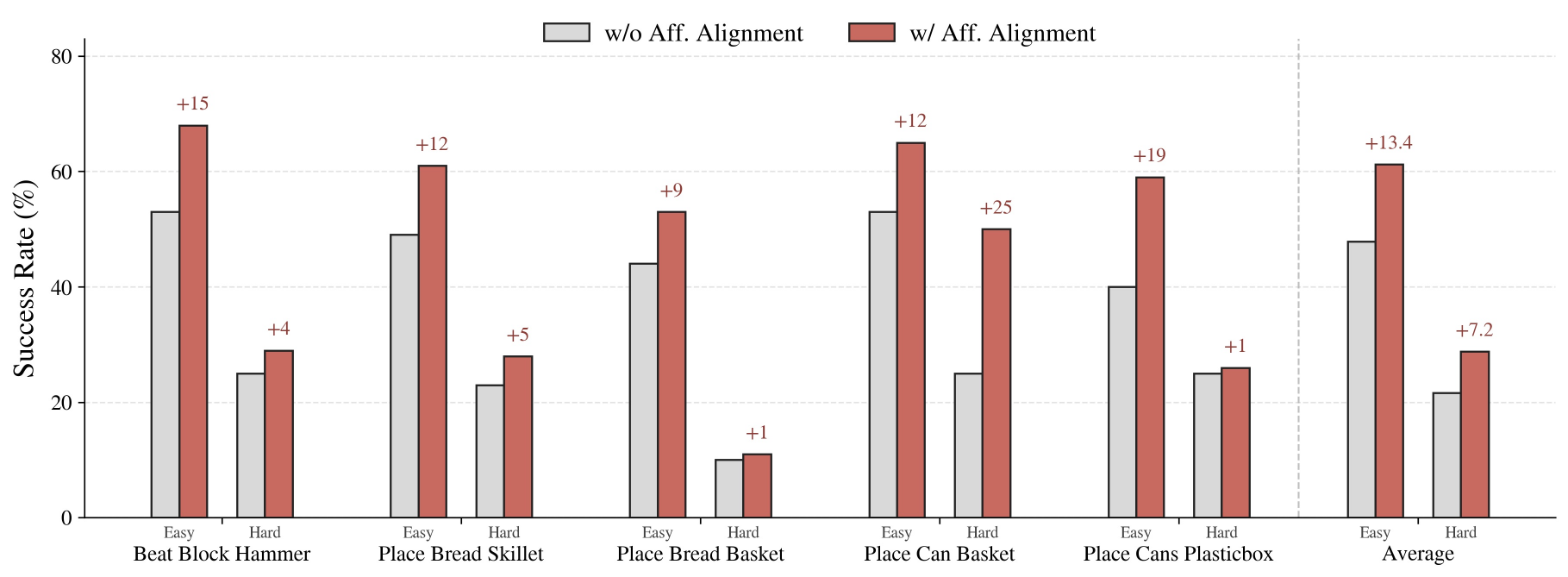} 
    \caption{\textbf{Ablation study of the implicit affordance representation alignment mechanism.} We compare the manipulation success rates on five RoboTwin2.0 simulation tasks with and without implicit affordance representation alignment. The results show that introducing this alignment mechanism significantly improves manipulation performance, especially in visually distracting scenarios.}
    \label{fig:vla_framework}
\end{figure*}

\begin{figure}[t]
    \centering
    \includegraphics[width=\columnwidth]{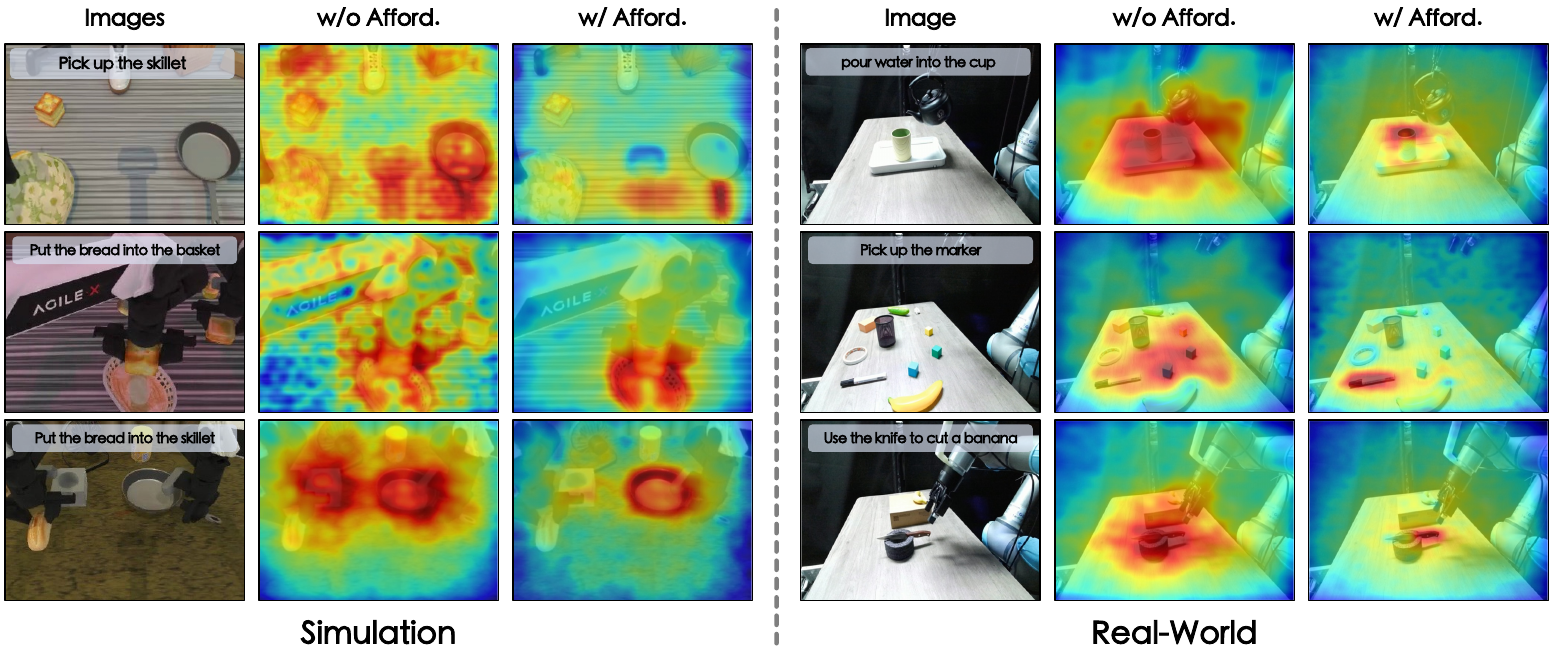}
    \caption{\textbf{Visualization of visual attention}. We visualize the attention heatmaps of the 12th-layer VLA visual representations. The results show that, after alignment, the model shifts its visual attention from background or non-functional regions to functional interaction regions relevant to task execution.}
    \label{fig:right_column_image}
\end{figure}

\begin{figure}[t]
    \centering
    \includegraphics[width=\columnwidth]{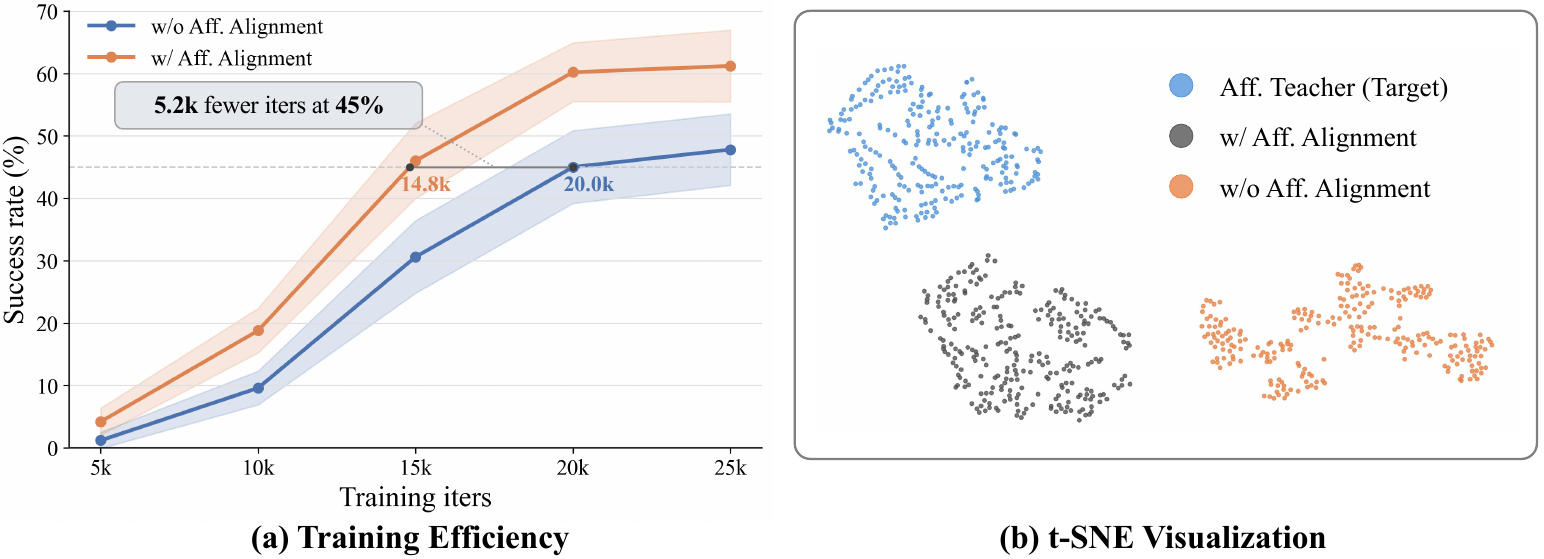}
    \caption{(a) \textbf{Training efficiency comparison.} We compare training efficiency on RoboTwin2.0 tasks with and without implicit affordance representation alignment. The results show that, to reach an average success rate of 45\%, the model with affordance representation alignment requires about 5.2k fewer training iterations. (b) \textbf{t-SNE visualization.} The aligned VLA visual representations show a distribution structure more similar to the affordance teacher representations while still preserving their own representation characteristics.}
    \label{fig:right_column_image}
\end{figure}

\textbf{Alignment Effect.} To evaluate whether implicit affordance representation alignment effectively improves the model’s performance, we compare AffordVLA on RoboTwin2.0 with and without affordance representation alignment, as shown in Fig.~8. The results show that introducing affordance representation alignment improves the success rate across multiple tasks, with similarly clear gains under the Hard setting involving complex backgrounds, object occlusions, and visual distractions. This indicates that the supervision provided by the external affordance teacher can effectively guide the VLA intermediate visual representations from global appearance cues toward task-relevant functional interaction regions, thereby improving the stability and robustness of the policy in complex manipulation scenarios. In contrast, without the alignment loss, the model is more easily affected by background textures, irrelevant objects, and non-functional regions, leading to unstable target localization and action execution failures. These results confirm that affordance representation alignment is a key source of the performance improvement of AffordVLA.

\textbf{Attention Visualization.} To further analyze the reason for the performance improvement, we visualize the attention heatmaps of the intermediate visual layer of the VLA with and without affordance representation alignment, as shown in Fig.~9. Without alignment, the model’s attention is often dispersed across the entire object, background regions, or other distractors, making it difficult to consistently focus on the functional parts directly related to task execution. In contrast, after introducing affordance representation alignment, the model’s attention becomes more concentrated on task-relevant regions, such as tool handles, container openings, target objects, and contact regions. This phenomenon shows that affordance representation alignment not only improves numerical performance but also reshapes the attention pattern of VLA intermediate visual representations, endowing the model with stronger manipulation-centric perception.

\textbf{Training Efficiency.} Fig.~10(a) shows the difference in training efficiency with and without affordance representation alignment. Under the same number of training steps, the model with alignment supervision usually reaches higher success rates more quickly. From the perspective of the number of training steps required to reach the same success rate, introducing affordance representation alignment clearly accelerates convergence. This suggests that the intermediate representation supervision provided by the affordance teacher offers a more direct visual learning signal for the VLA, enabling the model to identify task-relevant functional regions more efficiently. These results show that implicit representation alignment effectively improves both training efficiency and data efficiency.

\textbf{Representation Analysis.} To evaluate the effectiveness of the alignment mechanism at the feature-space level, we further apply t-SNE~\cite{t-sne} to visualize VLA intermediate visual representations, as shown in Fig.~10(b). The visualization shows that, after alignment, the distribution structure of VLA visual representations becomes more similar to that of the affordance teacher representations, while their cluster centers remain distinct from the teacher representations. This indicates that, during alignment, VLA visual representations internalize task-conditioned affordance priors, while still preserving the semantic information specific to the VLA modality.

\section{Conclusion}
We propose AffordVLA, a framework that injects affordance perception into VLA visual representations through implicit representation alignment. By introducing a frozen zero-shot affordance teacher, AffordVLA internalizes task-conditioned affordance priors into VLA intermediate visual representations during training, while preserving the original end-to-end inference structure without requiring additional affordance modules or explicit affordance inputs. Extensive simulation and real-world robot experiments show that AffordVLA effectively improves manipulation performance and robustness in visually distracting and unstructured environments, while maintaining efficient inference.

Despite the promising results, AffordVLA still has several limitations. First, its training relies on the affordance teacher to provide representation supervision, and thus teacher noise in complex scenarios may affect the learning process. Second, the current method mainly injects affordance information into the intermediate visual representations of the VLA, without directly involving the action generation process. Therefore, it may still be limited in tasks that require precise contact or continuous action adjustment. Finally, its generalization ability to longer-horizon tasks, multi-subgoal tasks, and different robotic platforms remains to be further validated.

In future work, we will further improve AffordVLA along these directions. Specifically, we aim to use affordance visual representations to support task decomposition and stage transitions, enabling robots to execute more complex long-horizon and multi-stage manipulation tasks. We will also explore how to more tightly integrate affordance priors with the action generation process, and further evaluate AffordVLA on more real-world robotic platforms and diverse scenarios.



\bibliographystyle{IEEEtran}
\bibliography{reference/myreference}

@IEEEtranBSTCTL{IEEEexample:BSTcontrol,
  CTLuse_forced_etal       = "yes",
  CTLmax_names_forced_etal = "6",
  CTLnames_show_etal       = "3"
}

@article{dong-2,
  title={Flexible robotic hand harnesses large deformations for full-coverage human-like multimodal haptic perception},
  author={Wang, Yanzhe and Guo, Haotian and Wu, Hao and Dong, Huixu},
  journal={Nature Communications},
  year={2025},
  publisher={Nature Publishing Group UK London}
}

@inproceedings{3dapnet,
  title={Language-conditioned affordance-pose detection in 3D point clouds},
  author={Nguyen, Toan and Vu, Minh Nhat and Huang, Baoru and Van Vo, Tuan and Truong, Vy and Le, Ngan and Vo, Thieu and Le, Bac and Nguyen, Anh},
  booktitle={Proc. IEEE Int. Conf. Robot. Autom.},
  pages={3071--3078},
  address={Yokohama, Japan},
  year={2024}
}

@inproceedings{uad,
  title={Uad: Unsupervised affordance distillation for generalization in robotic manipulation},
  author={Tang, Yihe and Huang, Wenlong and Wang, Yingke and Li, Chengshu and Yuan, Roy and Zhang, Ruohan and Wu, Jiajun and Fei-Fei, Li},
  booktitle={Proc. IEEE Int. Conf. Robot. Autom.},
  pages={3822--3831},
  address={Atlanta, GA, USA},
  year={2025}
}

@inproceedings{affordancenet,
  title={Affordancenet: An end-to-end deep learning approach for object affordance detection},
  author={Do, Thanh-Toan and Nguyen, Anh and Reid, Ian},
  booktitle={Proc. IEEE Int. Conf. Robot. Autom.},
  pages={5882--5889},
  address={Brisbane, Australia},
  year={2018}
}

@inproceedings{mcr,
  title={Robots Pre-train Robots: Manipulation-Centric Robotic Representation from Large-Scale Robot Datasets},
  author={Guangqi Jiang and Yifei Sun and Tao Huang and others},
  booktitle={Proc. 13th Int. Conf. Learn. Represent.},
  address={Singapore},
  year={2025}
}

@inproceedings{manipvqa,
  title={Manipvqa: Injecting robotic affordance and physically grounded information into multi-modal large language models},
  author={Huang, Siyuan and Ponomarenko, Iaroslav and Jiang, Zhengkai and Li, Xiaoqi and Hu, Xiaobin and Gao, Peng and Li, Hongsheng and Dong, Hao},
  booktitle={Proc. IEEE/RSJ Int. Conf. Intell. Robots Syst.},
  pages={7580--7587},
  address={Abu Dhabi, United Arab Emirates},
  year={2024}
}

@article{tars,
  title={TARS: Tactile affordance in robot synesthesia for dexterous manipulation},
  author={Wu, Qiwei and Wang, Haidong and Zhou, Jiayu and Xiong, Xiaogang and Lou, Yunjiang},
  journal={IEEE Robotics and Automation Letters},
  volume={10},
  number={1},
  pages={327--334},
  year={2024},
  publisher={IEEE}
}

@inproceedings{omnimanip,
  title={Omnimanip: Towards general robotic manipulation via object-centric interaction primitives as spatial constraints},
  author={Pan, Mingjie and Zhang, Jiyao and Wu, Tianshu and Zhao, Yinghao and Gao, Wenlong and Dong, Hao},
  booktitle={Proc. IEEE/CVF Conf. Comput. Vis. Pattern Recognit.},
  pages={17359--17369},
  address={Nashville, TN, USA},
  year={2025}
}

@inproceedings{a0,
  title={A0: An affordance-aware hierarchical model for general robotic manipulation},
  author={Xu, Rongtao and Zhang, Jian and Guo, Minghao and Wen, Youpeng and Yang, Haoting and Lin, Min and Huang, Jianzheng and Li, Zhe and Zhang, Kaidong and Wang, Liqiong and others},
  booktitle={Proc. IEEE/CVF Int. Conf. Comput. Vis.},
  pages={13491--13501},
  address={Honolulu, HI, USA},
  year={2025}
}

@article{sa-dem,
  title={SA-DEM: Dexterous Extrinsic Robotic Manipulation of Non-Graspable Objects via Stiffness-Aware Dual-Stage Reinforcement Learning},
  author={Wang, Yanzhe and Yu, Wei and Wu, Hao and Guo, Haotian and Dong, Huixu},
  journal={IEEE Transactions on Automation Science and Engineering},
  volume={23},
  pages={347--362},
  year={2025},
  publisher={IEEE}
}

@inproceedings{rt-1,
  title={RT-1: Robotics Transformer for Real-World Control at Scale},
  author={Brohan, Anthony and others},
  booktitle={Proc. Robot. Sci. Syst.},
  address={Daegu, Republic of Korea},
  year={2023}
}

@inproceedings{rt-2,
  title={Rt-2: Vision-language-action models transfer web knowledge to robotic control},
  author={Zitkovich, Brianna and Yu, Tianhe and Xu, Sichun and Xu, Peng and Xiao, Ted and Xia, Fei and Wu, Jialin and Wohlhart, Paul and Welker, Stefan and Wahid, Ayzaan and others},
  booktitle={Proc. Conf. Robot Learn.},
  pages={2165--2183},
  address={Atlanta, GA, USA},
  year={2023}
}

@article{pi_0,
  title={$\pi_0$: A Vision-Language-Action Flow Model for General Robot Control},
  author={Black, Kevin and Brown, Noah and others},
  journal={arXiv preprint arXiv:2410.24164},
  year={2024}
}

@article{pi_05,
  title={$\pi_{0.5}$: a Vision-Language-Action Model with Open-World Generalization},
  author={Intelligence, Physical and Black, Kevin and Brown, Noah and Darpinian, James and Dhabalia, Karan and Driess, Danny and Esmail, Adnan and Equi, Michael and Finn, Chelsea and Fusai, Niccolo and others},
  journal={arXiv preprint arXiv:2504.16054},
  year={2025}
}

@inproceedings{openvla,
  title={Open{VLA}: An Open-Source Vision-Language-Action Model},
  author={Moo Jin Kim and Karl Pertsch and Siddharth Karamcheti and Ted Xiao and Ashwin Balakrishna and Suraj Nair and Rafael Rafailov and Ethan P Foster and Pannag R Sanketi and Quan Vuong and Thomas Kollar and Benjamin Burchfiel and Russ Tedrake and Dorsa Sadigh and Sergey Levine and Percy Liang and Chelsea Finn},
  booktitle={Proc. 8th Conf. Robot Learn.},
  address={Munich, Germany},
  year={2024},
}

@article{gr00t-n1,
  title={Gr00t n1: An open foundation model for generalist humanoid robots},
  author={Bjorck, Johan and Casta{\~n}eda, Fernando and Cherniadev, Nikita and Da, Xingye and Ding, Runyu and Fan, Linxi and Fang, Yu and Fox, Dieter and Hu, Fengyuan and Huang, Spencer and others},
  journal={arXiv preprint arXiv:2503.14734},
  year={2025}
}

@article{vla-jepa,
  title={VLA-JEPA: Enhancing Vision-Language-Action Model with Latent World Model},
  author={Sun, Jingwen and Zhang, Wenyao and Qi, Zekun and Ren, Shaojie and Liu, Zezhi and Zhu, Hanxin and Sun, Guangzhong and Jin, Xin and Chen, Zhibo},
  journal={arXiv preprint arXiv:2602.10098},
  year={2026}
}

@inproceedings{reconvla,
  title={Reconvla: Reconstructive vision-language-action model as effective robot perceiver},
  author={Song, Wenxuan and Zhou, Ziyang and Zhao, Han and Chen, Jiayi and Ding, Pengxiang and Yan, Haodong and Huang, Yuxin and Tang, Feilong and Wang, Donglin and Li, Haoang},
  booktitle={Proc. AAAI Conf. Artif. Intell.},
  volume={40},
  number={22},
  pages={18549--18557},
  address={Singapore},
  year={2026}
}

@inproceedings{sf,
  title={Spatial Forcing: Implicit Spatial Representation Alignment for Vision-language-action Model},
  author={Fuhao Li and Wenxuan Song and Han Zhao and Jingbo Wang and Pengxiang Ding and Donglin Wang and Long ZENG and Haoang Li},
  booktitle={Proc. 14th Int. Conf. Learn. Represent.},
  address={Rio de Janeiro, Brazil},
  year={2026}
}

@inproceedings{rt-affordance,
  title={Rt-affordance: Affordances are versatile intermediate representations for robot manipulation},
  author={Nasiriany, Soroush and Kirmani, Sean and Ding, Tianli and Smith, Laura and Zhu, Yuke and Driess, Danny and Sadigh, Dorsa and Xiao, Ted},
  booktitle={Proc. IEEE Int. Conf. Robot. Autom.},
  pages={8249--8257},
  address={Atlanta, GA, USA},
  year={2025}
}

@inproceedings{moka,
  title={Moka: Open-world robotic manipulation through mark-based visual prompting},
  author={Fang, Kuan and Liu, Fangchen and Abbeel, Pieter and Levine, Sergey},
  booktitle={Proc. Robot. Sci. Syst.},
  address={Delft, Netherlands},
  year={2024}
}

@article{kbag-net,
  title={Knowledge enhanced bottom-up affordance grounding for robotic interaction},
  author={Qu, Wen and others},
  journal={PeerJ Computer Science},
  volume={10},
  pages={e2097},
  year={2024},
  publisher={PeerJ Inc.}
}

@inproceedings{u-sst,
  title={Uncertainty-aware state space transformer for egocentric 3d hand trajectory forecasting},
  author={Bao, Wentao and Chen, Lele and others},
  booktitle={Proc. IEEE/CVF Int. Conf. Comput. Vis.},
  pages={13702--13711},
  address={Paris, France},
  year={2023}
}

@article{vla-survey,
  title={Vision language action models in robotic manipulation: A systematic review},
  author={Din, Muhayy Ud and Akram, Waseem and Saoud, Lyes Saad and Rosell, Jan and Hussain, Irfan},
  journal={arXiv preprint arXiv:2507.10672},
  year={2025}
}

@inproceedings{coa-vla,
  title={Coa-vla: Improving vision-language-action models via visual-text chain-of-affordance},
  author={Li, Jinming and Zhu, Yichen and Tang, Zhibin and Wen, Junjie and Zhu, Minjie and Liu, Xiaoyu and Li, Chengmeng and Cheng, Ran and Peng, Yaxin and Peng, Yan and others},
  booktitle={Proc. IEEE/CVF Int. Conf. Comput. Vis.},
  pages={9759--9769},
  address={Honolulu, HI, USA},
  year={2025}
}

@article{palm,
  title={PALM: Progress-Aware Policy Learning via Affordance Reasoning for Long-Horizon Robotic Manipulation},
  author={Liu, Yuanzhe and Zhu, Jingyuan and Mo, Yuchen and Li, Gen and Cao, Xu and Jin, Jin and Shen, Yifan and Li, Zhengyuan and Yu, Tianjiao and Yuan, Wenzhen and others},
  journal={arXiv preprint arXiv:2601.07060},
  year={2026}
}

@article{gpt-4,
  title={Gpt-4 technical report},
  author={Achiam, Josh and Adler, Steven and Agarwal, Sandhini and Ahmad, Lama and Akkaya, Ilge and Aleman, Florencia Leoni and Almeida, Diogo and Altenschmidt, Janko and Altman, Sam and Anadkat, Shyamal and others},
  journal={arXiv preprint arXiv:2303.08774},
  year={2023}
}

@article{qwen3,
  title={Qwen3-vl technical report},
  author={Bai, Shuai and Cai, Yuxuan and Chen, Ruizhe and Chen, Keqin and Chen, Xionghui and Cheng, Zesen and Deng, Lianghao and Ding, Wei and Gao, Chang and Ge, Chunjiang and others},
  journal={arXiv preprint arXiv:2511.21631},
  year={2025}
}

@inproceedings{rdt,
  title={{RDT}-1B: a Diffusion Foundation Model for Bimanual Manipulation},
  author={Songming Liu and Lingxuan Wu and Bangguo Li and Hengkai Tan and Huayu Chen and Zhengyi Wang and Ke Xu and Hang Su and Jun Zhu},
  booktitle={Proc. 13th Int. Conf. Learn. Represent.},
  address={Singapore},
  year={2025},
}

@inproceedings{oxe,
  title={Open x-embodiment: Robotic learning datasets and rt-x models: Open x-embodiment collaboration 0},
  author={O’Neill, Abby and Rehman, Abdul and Maddukuri, Abhiram and Gupta, Abhishek and Padalkar, Abhishek and Lee, Abraham and Pooley, Acorn and Gupta, Agrim and Mandlekar, Ajay and Jain, Ajinkya and others},
  booktitle={Proc. IEEE Int. Conf. Robot. Autom.},
  pages={6892--6903},
  address={Yokohama, Japan},
  year={2024}
}

@inproceedings{cot-vla,
  title={Cot-vla: Visual chain-of-thought reasoning for vision-language-action models},
  author={Zhao, Qingqing and Lu, Yao and others},
  booktitle={Proc. IEEE/CVF Conf. Comput. Vis. Pattern Recognit.},
  pages={1702--1713},
  address={Nashville, TN, USA},
  year={2025}
}

@article{lare-vla,
  title={Latent Reasoning VLA: Latent Thinking and Prediction for Vision-Language-Action Models},
  author={Bai, Shuanghao and Lyu, Jing and Zhou, Wanqi and Li, Zhe and Wang, Dakai and Xing, Lei and Zhao, Xiaoguang and Wang, Pengwei and Wang, Zhongyuan and Chi, Cheng and others},
  journal={arXiv preprint arXiv:2602.01166},
  year={2026}
}

@article{f1,
  title={F1: A vision-language-action model bridging understanding and generation to actions},
  author={Lv, Qi and Kong, Weijie and Li, Hao and Zeng, Jia and Qiu, Zherui and Qu, Delin and Song, Haoming and Chen, Qizhi and Deng, Xiang and Pang, Jiangmiao},
  journal={arXiv preprint arXiv:2509.06951},
  year={2025}
}

@article{vla-rl,
  title={Vla-rl: Towards masterful and general robotic manipulation with scalable reinforcement learning},
  author={Lu, Guanxing and Guo, Wenkai and Zhang, Chubin and Zhou, Yuheng and Jiang, Haonan and Gao, Zifeng and Tang, Yansong and Wang, Ziwei},
  journal={arXiv preprint arXiv:2505.18719},
  year={2025}
}

@article{pi_06*,
  title={$\pi^{*}_{0.6}$: a VLA That Learns From Experience},
  author={Intelligence, Physical and Amin, Ali and Aniceto, Raichelle and Balakrishna, Ashwin and Black, Kevin and Conley, Ken and Connors, Grace and Darpinian, James and Dhabalia, Karan and DiCarlo, Jared and others},
  journal={arXiv preprint arXiv:2511.14759},
  year={2025}
}

@article{ig-rft,
  title={IG-RFT: An Interaction-Guided RL Framework for VLA Models in Long-Horizon Robotic Manipulation},
  author={Su, Zhian and Kong, Weijie and Dong, Haonan and Dong, Huixu},
  journal={arXiv preprint arXiv:2602.20715},
  year={2026}
}

@article{Gibson,
  title={The Ecological Approach to Visual Perception},
  author={Gibson, James J},
  journal={Hilldale, USA},
  volume={1},
  number={2},
  pages={67--82},
  year={1977}
}

@article{affpose,
  title={AffPose: An Integrated RGB-Based Framework for Simultaneous Pose Estimation and Affordance Detection in Robotic Tool Manipulation},
  author={Kong, Weijie and Lin, Zhaohui and Yu, Wei and Guo, Haotian and Su, Zhian and Dong, Huixu},
  journal={IEEE Robotics and Automation Letters},
  year={2025},
  publisher={IEEE}
}

@inproceedings{affordancellm,
  title={Affordancellm: Grounding affordance from vision language models},
  author={Qian, Shengyi and Chen, Weifeng and Bai, Min and others},
  booktitle={Proc. IEEE/CVF Conf. Comput. Vis. Pattern Recognit.},
  pages={7587--7597},
  address={Seattle, WA, USA},
  year={2024}
}

@article{relanet,
  title={Object affordance detection with relationship-aware network},
  author={Zhao, Xue and Cao, Yang and Kang, Yu},
  journal={Neural Computing and Applications},
  volume={32},
  number={18},
  pages={14321--14333},
  year={2020},
  publisher={Springer}
}

@article{acpl,
  title={Affordance-centric policy learning: Sample efficient and generalisable robot policy learning using affordance-centric task frames},
  author={Rana, Krishan and Abou-Chakra, Jad and Garg, Sourav and Lee, Robert and Reid, Ian and Suenderhauf, Niko},
  journal={arXiv preprint arXiv:2410.12124},
  volume={2},
  year={2024},
  publisher={Oct}
}

@article{clover,
  title={Closed-loop visuomotor control with generative expectation for robotic manipulation},
  author={Bu, Qingwen and Zeng, Jia and Chen, Li and Yang, Yanchao and Zhou, Guyue and Yan, Junchi and Luo, Ping and Cui, Heming and Ma, Yi and Li, Hongyang},
  journal={Advances in Neural Information Processing Systems},
  volume={37},
  pages={139002--139029},
  year={2024}
}

@inproceedings{manipgpt,
  title={ManipGPT: Is Affordance Segmentation by Large Vision Models Enough for Articulated Object Manipulation?},
  author={Kim, Taewhan and Bae, Hojin and Li, Zeming and Li, Xiaoqi and Ponomarenko, Iaroslav and Wu, Ruihai and Dong, Hao},
  booktitle={Proc. IEEE/RSJ Int. Conf. Intell. Robots Syst.},
  pages={20974--20981},
  address={Hangzhou, China},
  year={2025}
}

@article{r3m,
  title={R3m: A universal visual representation for robot manipulation},
  author={Nair, Suraj and Rajeswaran, Aravind and Kumar, Vikash and Finn, Chelsea and Gupta, Abhinav},
  journal={arXiv preprint arXiv:2203.12601},
  year={2022}
}

@inproceedings{repa,
  title={Representation Alignment for Generation: Training Diffusion Transformers Is Easier Than You Think},
  author={Sihyun Yu and Sangkyung Kwak and Huiwon Jang and Jongheon Jeong and Jonathan Huang and Jinwoo Shin and Saining Xie},
  booktitle={Proc. 13th Int. Conf. Learn. Represent.},
  address={Singapore},
  year={2025}
}

@inproceedings{3drs,
  title={3DRS: MLLMs Need 3D-Aware Representation Supervision for Scene Understanding},
  author={Huang, Xiaohu and Wu, Jingjing and Xie, Qunyi and Han, Kai},
  booktitle={Adv. Neural Inf. Process. Syst.},
  address={San Diego, CA, USA},
  year={2025}
}

@inproceedings{genhancer,
  title={Genhancer: Imperfect generative models are secretly strong vision-centric enhancers},
  author={Ma, Shijie and Ge, Yuying and Wang, Teng and others},
  booktitle={Proc. IEEE/CVF Int. Conf. Comput. Vis.},
  pages={24402--24412},
  address={Honolulu, HI, USA},
  year={2025}
}

@inproceedings{ross,
  title={Reconstructive Visual Instruction Tuning},
  author={Haochen Wang and Anlin Zheng and Yucheng Zhao and Tiancai Wang and Zheng Ge and Xiangyu Zhang and Zhaoxiang Zhang},
  booktitle={Proc. 13th Int. Conf. Learn. Represent.},
  address={Singapore},
  year={2025}
}

@article{cma-vla,
  title={Cross-Modality Alignment Perception and Multi-Head Self-Attention Mechanism for Vision-Language-Action of Humanoid Robot},
  author={Ren, Bin and Shi, Diwei},
  journal={Sensors},
  volume={26},
  number={1},
  pages={165},
  year={2025},
  publisher={MDPI}
}

@inproceedings{spatialvla,
  title={Spatialvla: Exploring spatial representations for visual-language-action model},
  author={Qu, Delin and Song, Haoming and Chen, Qizhi and Yao, Yuanqi and Ye, Xinyi and Ding, Yan and Wang, Zhigang and Gu, JiaYuan and Zhao, Bin and Wang, Dong and others},
  booktitle={Proc. Robot. Sci. Syst.},
  address={Los Angeles, CA, USA},
  year={2025}
}

@article{flare,
  title={Flare: Robot learning with implicit world modeling},
  author={Zheng, Ruijie and Wang, Jing and Reed, Scott and Bjorck, Johan and Fang, Yu and Hu, Fengyuan and Jang, Joel and Kundalia, Kaushil and Lin, Zongyu and Magne, Loic and others},
  journal={arXiv preprint arXiv:2505.15659},
  year={2025}
}

@inproceedings{flow-matching,
  title={Flow Matching for Generative Modeling},
  author={Yaron Lipman and Ricky T. Q. Chen and Heli Ben-Hamu and Maximilian Nickel and Matthew Le},
  booktitle={Proc. 11th Int. Conf. Learn. Represent.},
  address={Kigali, Rwanda},
  year={2023}
}

@inproceedings{sam3,
  title={{SAM} 3: Segment Anything with Concepts},
  author={Nicolas Carion and Laura Gustafson and Yuan-Ting Hu and Shoubhik Debnath and Ronghang Hu and Didac Suris Coll-Vinent and Chaitanya Ryali and Kalyan Vasudev Alwala and Haitham Khedr and Andrew Huang and Jie Lei and Tengyu Ma and Baishan Guo and Arpit Kalla and Markus Marks and Joseph Greer and Meng Wang and Peize Sun and Roman R{\"a}dle and Triantafyllos Afouras and Effrosyni Mavroudi and Katherine Xu and Tsung-Han Wu and Yu Zhou and Liliane Momeni and RISHI HAZRA and Shuangrui Ding and Sagar Vaze and Francois Porcher and Feng Li and Siyuan Li and Aishwarya Kamath and Ho Kei Cheng and Piotr Dollar and Nikhila Ravi and Kate Saenko and Pengchuan Zhang and Christoph Feichtenhofer},
  booktitle={Proc. 14th Int. Conf. Learn. Represent.},
  address={Rio de Janeiro, Brazil},
  year={2026},
}

@inproceedings{huang2024,
  title={Deciphering Cross-Modal Alignment in Large Vision-Language Models via Modality Integration Rate},
  author={Huang, Qidong and others},
  booktitle={Proc. IEEE/CVF Int. Conf. Comput. Vis.},
  pages={218--227},
  address={Honolulu, HI, USA},
  year={2025}
}

@inproceedings{3doi,
  title={Understanding 3d object interaction from a single image},
  author={Qian, Shengyi and Fouhey, David F},
  booktitle={Proc. IEEE/CVF Int. Conf. Comput. Vis.},
  pages={21753--21763},
  address={Paris, France},
  year={2023}
}

@inproceedings{ooal,
  title={One-shot open affordance learning with foundation models},
  author={Li, Gen and Sun, Deqing and Sevilla-Lara, Laura and Jampani, Varun},
  booktitle={Proc. IEEE/CVF Conf. Comput. Vis. Pattern Recognit.},
  pages={3086--3096},
  address={Seattle, WA, USA},
  year={2024}
}

@article{affordancesam,
  title={AffordanceSAM: Segment Anything Once More in Affordance Grounding},
  author={Jiang, Dengyang and Wang, Zanyi and Li, Hengzhuang and Dang, Sizhe and Ma, Teli and Wei, Wei and Dai, Guang and Zhang, Lei and Wang, Mengmeng},
  journal={arXiv preprint arXiv:2504.15650},
  year={2025}
}

@article{affogato,
  title={Affogato: Learning Open-Vocabulary Affordance Grounding with Automated Data Generation at Scale},
  author={Lee, Junha and Park, Eunha and Park, Chunghyun and Kang, Dahyun and Cho, Minsu},
  journal={arXiv preprint arXiv:2506.12009},
  year={2025}
}

@inproceedings{lisa,
  title={Lisa: Reasoning segmentation via large language model},
  author={Lai, Xin and Tian, Zhuotao and Chen, Yukang and Li, Yanwei and Yuan, Yuhui and Liu, Shu and Jia, Jiaya},
  booktitle={Proc. IEEE/CVF Conf. Comput. Vis. Pattern Recognit.},
  pages={9579--9589},
  address={Seattle, WA, USA},
  year={2024}
}

@inproceedings{mmsa,
  title={{MMR}: A Large-scale Benchmark Dataset for Multi-target and Multi-granularity Reasoning Segmentation},
  author={Donggon Jang and Yucheol Cho and Suin Lee and others},
  booktitle={Proc. 13th Int. Conf. Learn. Represent.},
  address={Singapore},
  year={2025}
}

@inproceedings{BiT-Align,
  title={Resource-Efficient Affordance Grounding with Complementary Depth and Semantic Prompts},
  author={Huang, Yizhou and others},
  booktitle={Proc. IEEE/RSJ Int. Conf. Intell. Robots Syst.},
  pages={7788--7795},
  address={Hangzhou, China},
  year={2025}
}

@inproceedings{r-mamba,
  title={Reasoning mamba: Hypergraph-guided region relation calculating for weakly supervised affordance grounding},
  author={Wang, Yuxuan and Wu, Aming and Yang, Muli and Min, Yukuan and Zhu, Yihang and Deng, Cheng},
  booktitle={Proc. IEEE/CVF Conf. Comput. Vis. Pattern Recognit.},
  pages={27618--27627},
  address={Nashville, TN, USA},
  year={2025}
}

@inproceedings{WSAG-PLSP,
  title={Weakly-Supervised Affordance Grounding Guided by Part-Level Semantic Priors},
  author={Peiran Xu and Yadong MU},
  booktitle={Proc. 13th Int. Conf. Learn. Represent.},
  address={Singapore},
  year={2025}
}

@inproceedings{WSMA,
  title={Weakly supervised multimodal affordance grounding for egocentric images},
  author={Xu, Lingjing and Gao, Yang and Song, Wenfeng and Hao, Aimin},
  booktitle={Proc. AAAI Conf. Artif. Intell.},
  volume={38},
  number={6},
  pages={6324--6332},
  address={Vancouver, Canada},
  year={2024}
}

@inproceedings{intra,
  title={Intra: Interaction relationship-aware weakly supervised affordance grounding},
  author={Jang, Ji Ha and Seo, Hoigi and Chun, Se Young},
  booktitle={Proc. Eur. Conf. Comput. Vis.},
  pages={18--34},
  address={Milan, Italy},
  year={2024}
}

@inproceedings{locate,
  title={Locate: Localize and transfer object parts for weakly supervised affordance grounding},
  author={Li, Gen and Jampani, Varun and Sun, Deqing and Sevilla-Lara, Laura},
  booktitle={Proc. IEEE/CVF Conf. Comput. Vis. Pattern Recognit.},
  pages={10922--10931},
  address={Vancouver, Canada},
  year={2023}
}

@inproceedings{AffCorrs,
  title={One-shot transfer of affordance regions? affcorrs!},
  author={Hadjivelichkov, Denis and Zwane, Sicelukwanda and Agapito, Lourdes and Deisenroth, Marc Peter and Kanoulas, Dimitrios},
  booktitle={Proc. Conf. Robot Learn.},
  pages={550--560},
  address={Atlanta, GA, USA},
  year={2023}
}

@inproceedings{agd20k,
  title={Learning affordance grounding from exocentric images},
  author={Luo, Hongchen and Zhai, Wei and Zhang, Jing and Cao, Yang and Tao, Dacheng},
  booktitle={Proc. IEEE/CVF Conf. Comput. Vis. Pattern Recognit.},
  pages={2252--2261},
  address={New Orleans, LA, USA},
  year={2022}
}

@inproceedings{InteractionHotspots,
  title={Grounded human-object interaction hotspots from video},
  author={Nagarajan, Tushar and Feichtenhofer, Christoph and Grauman, Kristen},
  booktitle={Proc. IEEE/CVF Int. Conf. Comput. Vis.},
  pages={8688--8697},
  address={Seoul, Republic of Korea},
  year={2019}
}

@article{dp,
  title={Diffusion policy: Visuomotor policy learning via action diffusion},
  author={Chi, Cheng and Xu, Zhenjia and Feng, Siyuan and Cousineau, Eric and Du, Yilun and Burchfiel, Benjamin and Tedrake, Russ and Song, Shuran},
  journal={The International Journal of Robotics Research},
  volume={44},
  number={10-11},
  pages={1684--1704},
  year={2025},
  publisher={Sage Publications Sage UK: London, England}
}

@inproceedings{dp3,
  title={3D Diffusion Policy: Generalizable Visuomotor Policy Learning via Simple 3D Representations},
  author={Ze, Yanjie and Zhang, Gu and Zhang, Kangning and Hu, Chenyuan and Wang, Muhan and Xu, Huazhe},
  booktitle={Proc. Robot. Sci. Syst.},
  address={Delft, Netherlands},
  year={2024}
}

@inproceedings{act,
  title={Learning Fine-Grained Bimanual Manipulation with Low-Cost Hardware},
  author={Zhao, Tony and Kumar, Vikash and Levine, Sergey and Finn, Chelsea},
  booktitle={Proc. Robot. Sci. Syst.},
  address={Daegu, Republic of Korea},
  year={2023}
}

@article{robotwin,
  title={Robotwin 2.0: A scalable data generator and benchmark with strong domain randomization for robust bimanual robotic manipulation},
  author={Chen, Tianxing and Chen, Zanxin and Chen, Baijun and Cai, Zijian and Liu, Yibin and Li, Zixuan and Liang, Qiwei and Lin, Xianliang and Ge, Yiheng and Gu, Zhenyu and others},
  journal={arXiv preprint arXiv:2506.18088},
  year={2025}
}

@article{kld,
  title={What do different evaluation metrics tell us about saliency models?},
  author={Bylinskii, Zoya and Judd, Tilke and Oliva, Aude and Torralba, Antonio and Durand, Fr{\'e}do},
  journal={IEEE transactions on pattern analysis and machine intelligence},
  volume={41},
  number={3},
  pages={740--757},
  year={2018},
  publisher={IEEE}
}

@article{sim,
  title={Color indexing},
  author={Swain, Michael J and Ballard, Dana H},
  journal={International journal of computer vision},
  volume={7},
  number={1},
  pages={11--32},
  year={1991},
  publisher={Springer}
}

@article{nss,
  title={Components of bottom-up gaze allocation in natural images},
  author={Peters, Robert J and others},
  journal={Vision research},
  volume={45},
  number={18},
  pages={2397--2416},
  year={2005},
  publisher={Elsevier}
}

@article{t-sne,
  title={Visualizing data using t-SNE.},
  author={Van der Maaten, Laurens and Hinton, Geoffrey},
  journal={Journal of machine learning research},
  volume={9},
  number={11},
  year={2008}
}

\vfill

\end{document}